\PassOptionsToPackage{xcdraw,table}{xcolor}

\documentclass[]{fairmeta}

\usepackage[utf8]{inputenc} 

\usepackage[style=apa,citestyle=authoryear-comp,dashed=false,isbn=false,backend=biber,maxcitenames=2,uniquename=false,uniquelist=false,eprint=true,natbib]{biblatex}

\usepackage{csquotes}
\usepackage[T1]{fontenc}    
\usepackage{hyperref}       
\usepackage{url}            
\usepackage{booktabs}       
\usepackage{doi}
\usepackage{amsfonts}       
\usepackage{nicefrac}       
\usepackage{microtype}      
\usepackage{graphicx}
\usepackage[export]{adjustbox}
\usepackage{caption}
\usepackage{subcaption}
\usepackage{xspace}
\usepackage{amsmath}
\usepackage{amssymb}
\usepackage{mathtools}
\usepackage{amsthm}
\usepackage[nameinlink]{cleveref}

\usepackage{xpatch}
\xpatchbibdriver{article}
  {\usebibmacro{doi+url}}
  {\usebibmacro{doi+url}%
   \newunit\newblock
   \usebibmacro{eprint}}
  {}
  {}

\xpatchbibdriver{misc}
  {\usebibmacro{doi+url}}
  {\usebibmacro{doi+url}%
   \newunit\newblock
   \usebibmacro{eprint}}
  {}
  {}

\DeclareFieldFormat{doi}{%
  \mkbibacro{DOI}\addcolon\space
  \ifhyperref
    {\href{https://doi.org/#1}{\nolinkurl{#1}}}
    {\nolinkurl{#1}}}

\title{Reassessing the Validity of Spurious Correlations Benchmarks}

\author[1]{Samuel J.~Bell}
\author[1,\dagger]{Diane Bouchacourt}
\author[1,\dagger]{Levent Sagun}

\affiliation[1]{Meta FAIR}

\contribution[\dagger]{Joint last author}

\abstract{Neural networks can fail when the data contains spurious correlations.
To understand this phenomenon, researchers have proposed numerous spurious correlations benchmarks upon which to evaluate mitigation methods.
However, we observe that these benchmarks exhibit substantial disagreement, with the best methods on one benchmark performing poorly on another. 
We explore this disagreement, and examine benchmark validity by defining three desiderata that a benchmark should satisfy in order to meaningfully evaluate methods. 
Our results have implications for both benchmarks and mitigations:
we find that certain benchmarks are not meaningful measures of method performance, and that several methods are not sufficiently robust for widespread use.
We present a simple recipe for practitioners to choose methods using the \emph{most similar} benchmark to their given problem.
}

\date{\today}
\correspondence{Samuel J.~Bell at \email{sjbell@meta.com}}

\newcommand{\Diane}[1]{{}}
\newcommand{\Dianeprop}[1]{{}}

\newcommand{\subpopbench}{\emph{SubpopBench}\xspace\gdef\subpopbench{SubpopBench\xspace}}
\newcommand{\citybirds}{\emph{Citybirds\xspace}\gdef\citybirds{Citybirds\xspace}}
\newcommand{\animalsvsplants}{\emph{Animals vs. Plants (AvP)\xspace}\gdef\animalsvsplants{AvP\xspace}}
\newcommand{\waterbirds}{\emph{Waterbirds\xspace}\gdef\waterbirds{Waterbirds\xspace}}
\newcommand{\celeba}{\emph{CelebA\xspace}\gdef\celeba{CelebA\xspace}}
\newcommand{\metashift}{\emph{MetaShift\xspace}\gdef\metashift{MetaShift\xspace}}
\newcommand{\nicopp}{\emph{NICO++\xspace}\gdef\nicopp{NICO++\xspace}}
\newcommand{\imagenetbg}{\emph{ImageNetBG\xspace}\gdef\imagenetbg{ImageNetBG\xspace}}
\newcommand{\chexpert}{\emph{CheXpert\xspace}\gdef\chexpert{CheXpert\xspace}}
\newcommand{\civilcomments}{\emph{CivilComments\xspace}\gdef\civilcomments{CivilComments\xspace}}
\newcommand{\multinli}{\emph{MultiNLI\xspace}\gdef\multinli{MultiNLI\xspace}}
\newcommand{\dollarstreet}{\emph{Dollar Street\xspace}\gdef\dollarstreet{Dollar Street\xspace}}

\newcommand{\ermfailure}{\emph{ERM Failure\xspace}\gdef\ermfailure{ERM Failure\xspace}}
\newcommand{\discpower}{\emph{Discriminative Power\xspace}\gdef\discpower{Discriminative Power\xspace}}
\newcommand{\convergentvalidity}{\emph{Convergent Validity\xspace}\gdef\convergentvalidity{Convergent Validity\xspace}}

\crefformat{section}{§#2#1#3}

\usepackage[disable,textsize=tiny]{todonotes}

\begin{document}

\maketitle

\section{Introduction}

\begin{figure}[h!]
    \centering
     \begin{subfigure}[b]{0.33\columnwidth}
        \includegraphics[trim={0 0 0 0},clip,width=\textwidth]{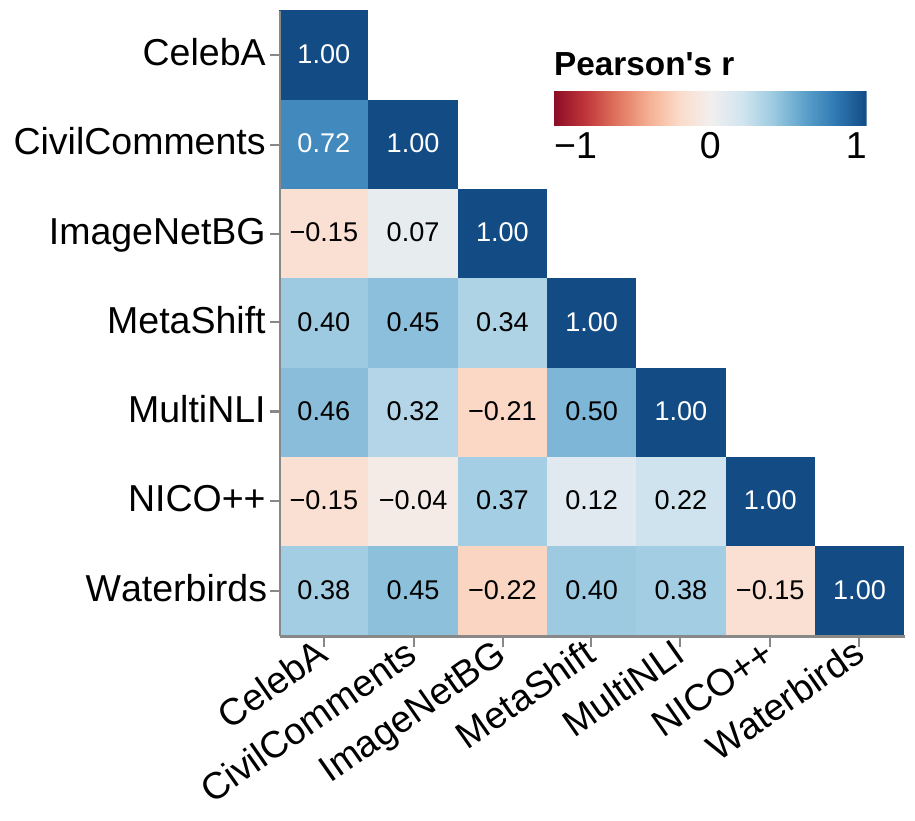}
        \caption{Benchmark agreement}
        \label{fig:published-results-sbp}
     \end{subfigure}
    \hspace{50pt}
    \begin{subfigure}[b]{0.209\textwidth}
        \centering
        \includegraphics[trim={0 0 0 0},clip,width=\textwidth]{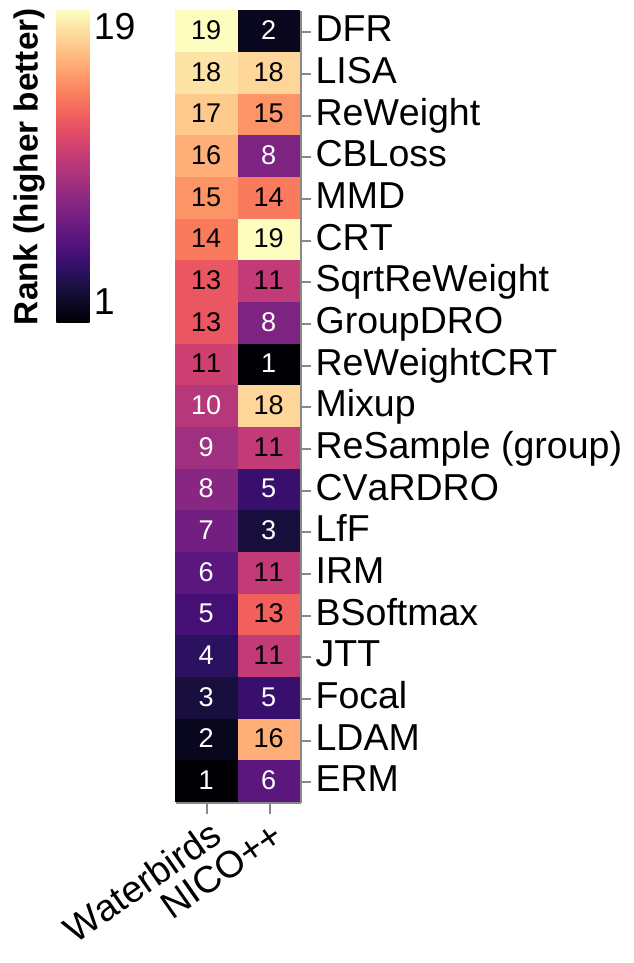}
        \caption{Method ranks}
        \label{fig:published-results-ranks}
    \end{subfigure}
    \caption{\textbf{Spurious correlations benchmarks disagree.} \textbf{(a)} Correlation between worst-group accuracies on different benchmarks reported by \citet{yang2023change}. \textbf{(b)} Waterbirds and NICO++ produce disagreeing ranks, such that the best method on Waterbirds (DFR) is the second worst on NICO++.}
    \label{fig:published-results}
\end{figure}

A striking failure mode of deep learning-based models is their susceptibility to spurious correlations, whereby models learn to use patterns that only hold in certain subsets of the data \citep{nagarajan2020understanding, geirhos2020shortcut}.
Researchers have produced numerous benchmarks for evaluating and comparing methods for mitigating spurious correlations, ultimately informing decisions as to which method is best.
In order to draw robust conclusions about which method to use, one would hope that different benchmarks produce similar results.
However, in \cref{fig:published-results} we observe this not to be the case: benchmarks often disagree, and methods that perform well on one benchmark perform poorly on others.

\newpage

Faced with multiple benchmarks, standard machine learning practice is to average over one's contradictory results.
Such averaging, however, neglects that different benchmarks may measure different things, only some of which correspond to the desired quality. 
This presents a key barrier to mitigating spurious correlations in practice.
When confronted by a new dataset exhibiting spurious correlations, which benchmark should one trust when deciding which method to apply?

In this work, we expose benchmark disagreement and analyze the validity of common spurious correlations benchmarks. 
To do this, we put forward a set of properties that a spurious correlations benchmark should satisfy, and introduce a model-dependent statistic that quantifies the benchmark's task difficulty due to spurious correlation. 
Using the established idea of convergent validity \citep{jacobs2021measurement}, we expect that two valid benchmarks testing similar things---exhibiting similar task difficulty due to spurious correlation---should rank methods similarly. 

\textbf{Our results reveal that certain benchmarks are not valid tools for evaluating mitigation method performance.}
Moreover, of all the methods evaluated here, only a small handful are robust to different tasks, while many exhibit strong performance only under specific conditions.
Finally, we provide an approach to translating between benchmark results and real-world datasets, using our model-dependent statistic to understand which benchmark is most relevant.

\subsection{Background}

Models trained with empirical risk minimization (ERM; \citealp[Ch.~1]{vapnik1999nature}) have a tendency to learn spurious correlations \citep{nagarajan2020understanding}, resulting in many real-world failures \citep{geirhos2020shortcut}. 
For example, chest x-ray classifiers latch onto physical features of the scanner that fail to generalize to new hospitals \citep{zech2018radiographs}.
In hate speech detection, models use dialect differences rather than learning the desired task, resulting in a disproportionate false-positive rate \citep{sap2019hate}.
Whether the result of sampling error or of harmful associations, spurious correlations  often lead to models which amplify bias and performance disparities \citep{zhao2017men, wang2019balanced}.

Many researchers have sought to understand why models rely on spurious correlations.
\citet{sagawa2020investigation} suggest that overparameterized models' bias against memorization leads them to rely on spurious correlations for minority samples. 
Given their well-known inductive bias towards simplicity \citep{kalimeris2019sgd, valle-perez2018deep, bell2023simplicity}, models use spurious correlations if they are simpler to learn than the intended function \citep{shah2020pitfalls, hall2022systematic}, where simplicity may be determined by model capacity \citep{sreekumar2023spurious}.
Given \emph{competing} and equally-informative spurious correlations, models will tend to pick the simplest option \citep{scimeca2021which}. 
Our effort to quantify task difficulty due to spurious correlation builds on these ideas of model-specific complexity. 

\citet{yang2023change} introduce \textit{\subpopbench} and evaluate 22 mitigation methods over benchmarks exhibiting different types of \emph{subpopulation shift}, including attribute and class imbalance and missing data.
Our work builds on \subpopbench due of its comprehensive set of methods and benchmarks, though we narrow our focus to only spurious correlations.
Several other benchmarking efforts exist, including those for evaluating spurious correlations mitigation \citep{joshi2023mitigating, lynch2023spawrious}, and other forms of subpopulation shift \citep{koh2021wilds, santurkar2020breeds, liu2023need}.
More broadly, \citet{gulrajani2020search} introduce the \emph{DomainBed} library for benchmarking performance in various domain generalization scenarios. 
Interestingly, both \citeauthor{gulrajani2020search} and \citeauthor{joshi2023mitigating} find that with sufficient hyperparameter tuning ERM can be surprisingly robust, motivating our consideration of ERM failure as a necessary benchmark property.

\begin{figure*}[tb!]
    \centering
    \begin{subfigure}[t]{0.22\textwidth}
        \centering
        \includegraphics[width=\textwidth,valign=t]{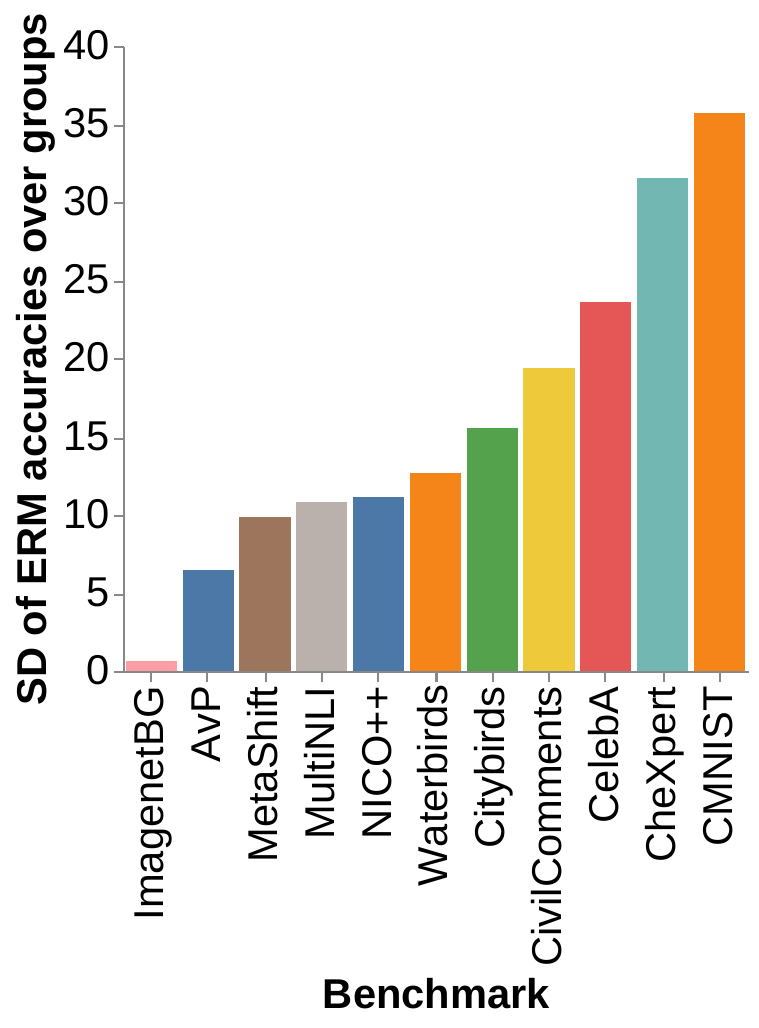}
        \caption{ERM variability}
        \label{fig:std-erm-std-method-erm}
    \end{subfigure}
    \hfill
    \begin{subfigure}[t]{0.22\textwidth}
        \centering
        \includegraphics[width=\textwidth,valign=t]{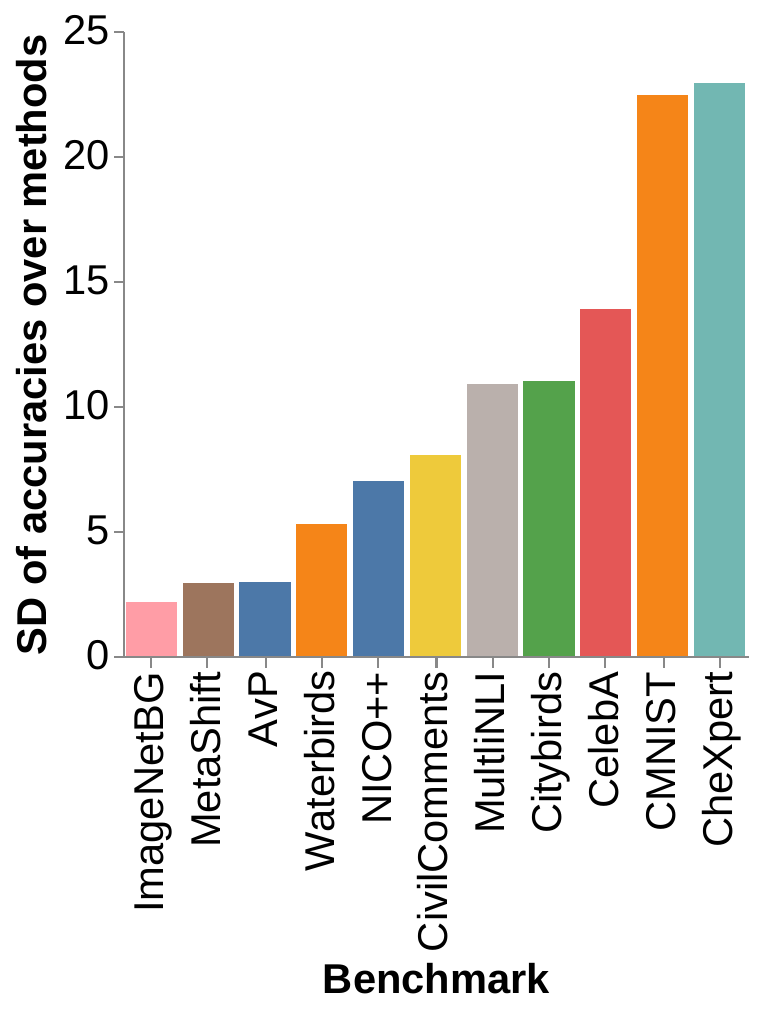}
        \vspace{1pt}
        \caption{Method variability}
        \label{fig:std-erm-std-method-method}
    \end{subfigure}
    \hfill
    \begin{subfigure}[t]{0.3252\textwidth}
        \centering
        \includegraphics[width=\textwidth,valign=t]{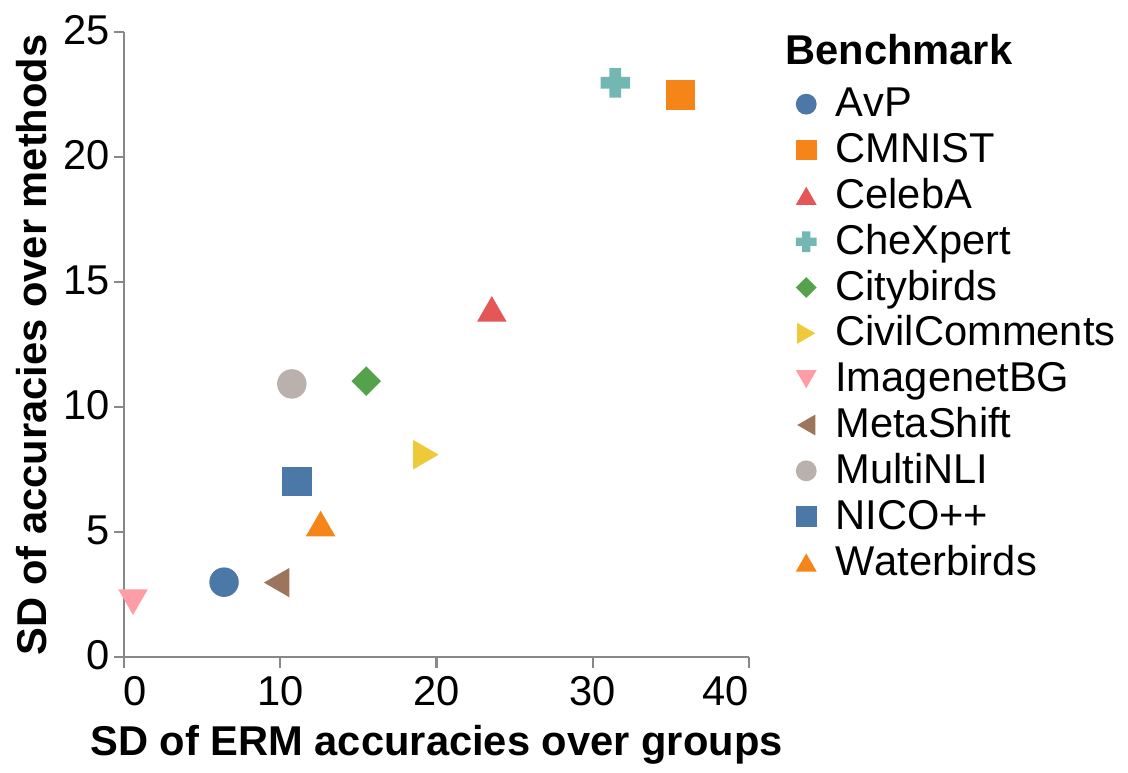}
        \vspace{34pt}
        \caption{ERM and method variability}
        \label{fig:std-erm-std-method-method-both}
    \end{subfigure}
    \caption{\textbf{(a)} Standard deviation (SD) of test accuracies over groups for an ERM-trained model. \textbf{(b)} SD of worst-group test accuracies over methods. \textbf{(c)} SD of ERM accuracies over groups vs. SD of worst-group test accuracies over methods. \textbf{Certain benchmarks, e.g.\ ImageNetBG, do not produce a ``worst group'', and result in tightly-clustered method performance.}}
    \label{fig:std-erm-std-method}
\end{figure*}

\subsection{Outline and Contributions}
\label{sec:outline}

We begin in \cref{sec:benchmark-agreement} with the simple observation that published results on spurious correlations benchmarks often disagree.
This motivates our investigation of \emph{what} benchmarks are measuring, and whether some benchmarks are more valid than others. 

In \cref{sec:benchmark-validity}, we introduce three desiderata---\ermfailure, \discpower, and \convergentvalidity---that capture key properties that a spurious correlations benchmark should satisfy in order to be a meaningful test of mitigation performance.
Our test for \convergentvalidity will call for some way of understanding precisely what a benchmark is actually measuring.
To address this, in \cref{sec:measuring-task-difficulty} we propose a model-dependent statistic that measures the difficulty of a task due to spurious correlation. 

In \cref{sec:method-robustness} we analyze the validity of eight benchmarks and the robustness of 22 mitigation methods. Finally, in \cref{sec:practical-recommendations} we present a recipe for mapping between benchmarks and real-world datasets, and evaluate our approach on geographically-diverse image classification. We conclude with the broader implications of our work in \cref{sec:discussion}.

\section{Not All Benchmarks Agree}
\label{sec:benchmark-agreement}

We define a benchmark as a pair of a task dataset (e.g.~an image classification task) and an evaluation metric for ranking methods (e.g.~worst-group test accuracy).
A spurious correlations benchmark is designed to measure how well methods can \emph{mitigate} the effects of spurious correlations.
These are typically designed so that conventional training yields poor performance on certain subsets of the data, referred to as \emph{groups}. 
If a benchmark's function is to allow us to conclude which methods are best, we would ideally like different benchmarks to agree with one other.

To our surprise, we find that they do not.
\Cref{fig:published-results-sbp} shows the (dis)agreement, as measured by Pearson's $r$, in worst-group test accuracies over benchmarks reported by \citet{yang2023change} (see \cref{sec:app-published-data}).
Two popular benchmarks, \emph{\waterbirds{}} and \emph{\celeba{}}, produce only mildly-correlated results, while results on \waterbirds{} and \emph{\nicopp{}} are negatively correlated.
This has a practical outcome: \cref{fig:published-results-ranks} shows that \textbf{the best} performing method on \waterbirds{} is \textbf{the second-worst} on \nicopp{}.
This disagreement is not due to benchmark saturation, as \waterbirds and \nicopp worst-group test accuracies have different means ($\bar{x}=78.4$ and $37.8$ respectively) but similarly large standard deviations ($s=6.2$ and $5.8$).
We report similar findings on WILDS \citep{koh2021wilds} (\cref{sec:app-published-data}).
This poses a challenge for the practitioner looking to mitigate spurious correlations: faced with inconsistent benchmarks, which method to use?

We propose a simple methodology to aid this problem by studying why benchmarks disagree, and evaluate whether some benchmarks are more valid measures of mitigation performance than others.
To do so, we define three desiderata that valid spurious correlations benchmarks should respect.

\section{Not All Benchmarks are Valid}
\label{sec:benchmark-validity}

For a benchmark to be a meaningful test of a method's ability to mitigate spurious correlations, we suggest it should satisfy three desiderata, which we call \emph{\ermfailure}, \emph{\discpower}, and \emph{\convergentvalidity}.

\textbf{\ermfailure} (\cref{sec:des-1-erm-failure}).
Spurious correlations mitigation methods are intended to prevent models from learning patterns that might perform well on average, but cause failures for certain groups. 
Thus, any benchmark intended to evaluate these methods should induce this problem when training with conventional empirical risk minimization (ERM), which by definition optimizes to minimize error averaged over all samples. 
\textbf{To satisfy \ermfailure, a benchmark should produce between-group performance disparities for models trained with ERM.}

\textbf{\discpower} (\cref{sec:des-2-disc-power}).
All benchmarks are intended for evaluation and comparison, and to support reasoning about \emph{which} methods are best. 
In order to serve this purpose, a benchmark must discriminate between methods and assign different scores to each.
These scores, in the spurious correlations setting, are typically worst-group test accuracies \citep{sagawa2020gdro}.
\textbf{To satisfy \discpower, a benchmark should produce different worst-group test accuracies for different methods.}

\begin{figure}[tb!]
    \captionsetup[subfigure]{justification=centering}
    \centering
    \begin{subfigure}[t]{0.16\textwidth}
        \centering
        \includegraphics[trim={0 0 0 0},clip,width=\textwidth]{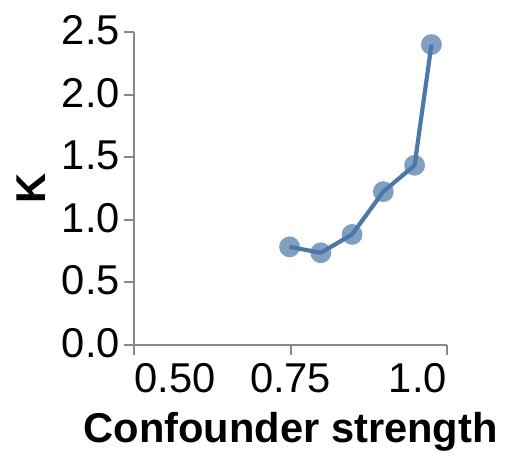}
        \caption{CelebA}
        \label{fig:bayes-factor-celeba-confounder}
    \end{subfigure}
    \begin{subfigure}[t]{0.16\textwidth}
        \centering
        \includegraphics[trim={0 0 0 0},clip,width=\textwidth]{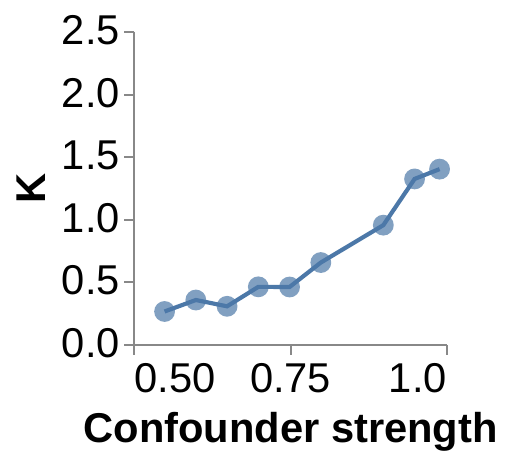}
        \caption{Waterbirds}
        \label{fig:bayes-factor-wb-confounder}
    \end{subfigure}
    \hfill
    \begin{subfigure}[t]{0.16\textwidth}
        \centering
        \includegraphics[trim={0 0 0 0},clip,width=\textwidth]{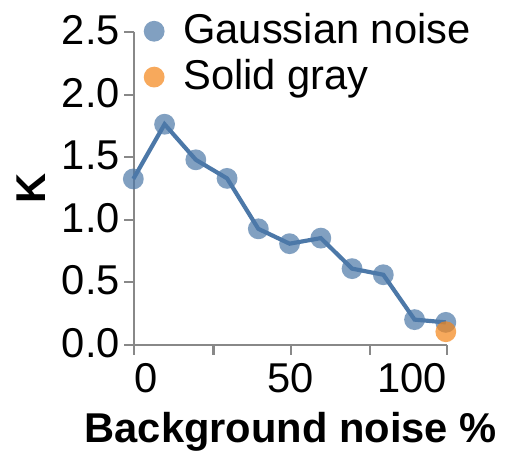}
        \caption{Waterbirds}
        \label{fig:bayes-factor-wb-bg-noise}
    \end{subfigure}
    \hfill
    \begin{subfigure}[t]{0.16\textwidth}
        \centering
        \includegraphics[trim={0 0 0 0},clip,width=\textwidth]{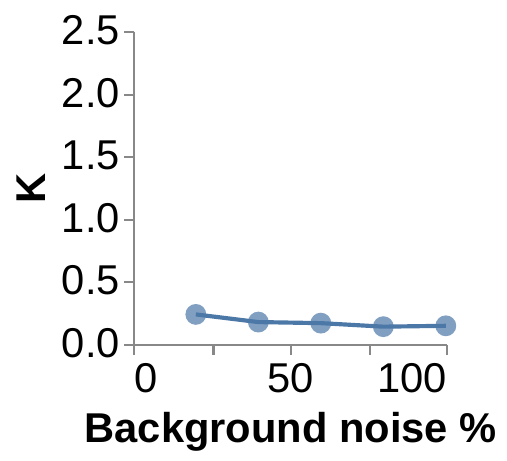}
        \caption{Waterbirds}
        \label{fig:bayes-factor-wb-bg-noise-no-corr}
    \end{subfigure}
    \hfill
    \begin{subfigure}[t]{0.16\textwidth}
        \centering
        \includegraphics[trim={0 0 0 0},clip,width=\textwidth]{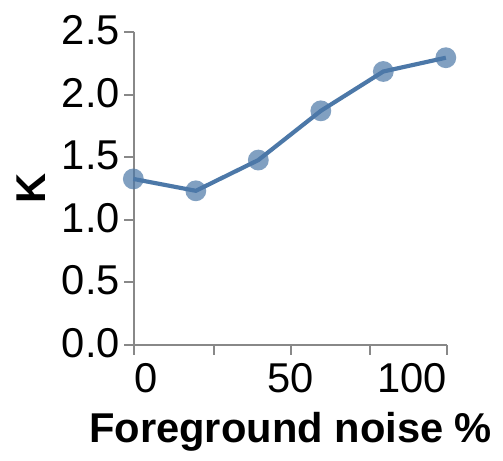}
        \caption{Waterbirds}
        \label{fig:bayes-factor-wb-fg-noise}
    \end{subfigure}
    \hfill
    \begin{subfigure}[t]{0.16\textwidth}
        \centering
        \includegraphics[trim={0 0 0 0},clip,width=\textwidth]{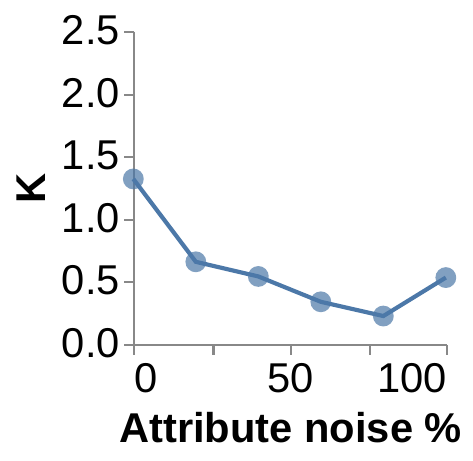}
        \caption{Waterbirds}
        \label{fig:bayes-factor-wb-attribute-noise}
    \end{subfigure}
    \caption{Task difficulty due to spurious correlation, as measured by Bayes Factor $K$, on modified benchmarks. Increasing the label-attribute correlation 
\textbf{(a, b)} and foreground noise \textbf{(e)} increases $K$, while increasing background noise \textbf{(c)} or applying a solid gray background \textbf{(c}, orange point\textbf{)} decreases $K$, except in the case where there is no correlation \textbf{(d)}. Attribute noise degrades the efficacy of $K$ \textbf{(f)}.}
    \label{fig:bayes-factor}
\end{figure}

\textbf{\convergentvalidity} (\cref{sec:des-3-convergent-validity}).
Even if a benchmark satisfies both \ermfailure and \discpower, it still needs to rank methods \emph{in a meaningful way}.
In other words, we want our benchmark to exhibit \emph{construct validity} \citep{jacobs2021measurement, blodgett2021stereotyping}, i.e.\ it should allow us to truly measure the extent to which methods mitigate spurious correlations.
Establishing construct validity of a benchmark is challenging without a ground truth, though one approach is to consider how the benchmark performs in relation to other benchmarks (that are themselves valid according to our first two desiderata).
\textbf{To satisfy \convergentvalidity, a benchmark should agree with other \emph{similar} benchmarks, and disagree with those that are \emph{dissimilar}.} 

\subsection{Evaluation Setup and Benchmarks}

We evaluate the validity of eight benchmarks included in \subpopbench:
\emph{\waterbirds{}} \citep{sagawa2020gdro}, \emph{\celeba{}} \citep{liu2015celeba}, 
\emph{\imagenetbg{}} (a.k.a. \emph{IN-9L Original}; \citealp{xiao2021imagenetbg}), \metashift{} (Indoor/Outdoor Cat vs. Dog; \citealp{liang2022metashift}), \emph{\nicopp{}} \citep{zhang2023nico}, \emph{CheXpert} \citep{irvin2019chexpert}, \emph{CivilComments} \citep{borkan2019civilcomments}, and \emph{MultiNLI} \citep{williams2018multinli}.

In addition, we develop two new benchmarks to sanity check our approach. \emph{Citybirds} is a clone of \waterbirds where confounding backgrounds are replaced with urban or rural scenes, such that \waterbirds and \citybirds should be equally valid. \emph{Animals vs. Plants (AvP)} is a binary image classification task of animals and plants from Asia and Europe. Geography is difficult to infer relative to class membership,\footnote{A linear classifier over pretrained representations failed to achieve better than chance performance on region identification, versus strong performance on the task. See \cref{sec:app-benchmark-datasets}.} so \animalsvsplants has no practical spurious correlation. 
Finally, as a real-world test case in \cref{sec:practical-recommendations}, we additionally evaluate on \dollarstreet{} \citep{rojas2022dollarstreet}, a multiclass classification task over household objects, where groups are geographic regions.
See \cref{sec:app-benchmark-datasets} for full details. 

We follow \citeauthor{yang2023change}'s methodology (see \cref{sec:app-subpopbench-evaluation}), report worst-group test accuracies, and perform model selection according to worst-group validation accuracies.

\subsection{Evaluating ERM Failure}
\label{sec:des-1-erm-failure}

We test for \ermfailure by evaluating the variability of per-group test accuracies of ERM-trained models.
High variability indicates that certain groups have worse performance than others,
whereas low variability indicates that there is no real ``worst group'', thus not satisfying the \ermfailure desideratum.

\Cref{fig:std-erm-std-method-erm} shows the standard deviation (SD) of per-group mean test accuracies for an ERM-trained model.
Immediately, we notice that \imagenetbg exhibits very low between-group variability and thus does not satisfy \ermfailure. \Diane{Can we make x axis bigger? To hard to read even zooming}

\begin{figure*}[tb!]
    \captionsetup[subfigure]{justification=centering}
    \centering
     \begin{subfigure}[b]{0.17\textwidth}
        \centering
        \includegraphics[trim={0 0 0 0},clip,width=\textwidth]{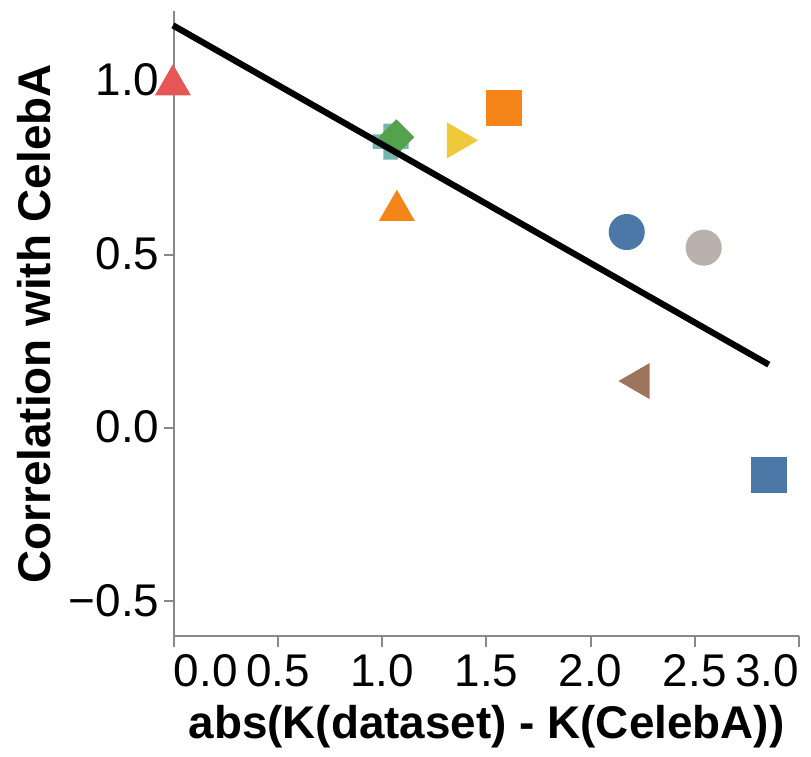}
         \caption{\celeba}
     \end{subfigure}
     \hfill
     \begin{subfigure}[b]{0.17\textwidth}
        \centering
        \includegraphics[trim={0 0 0 0},clip,width=\textwidth]{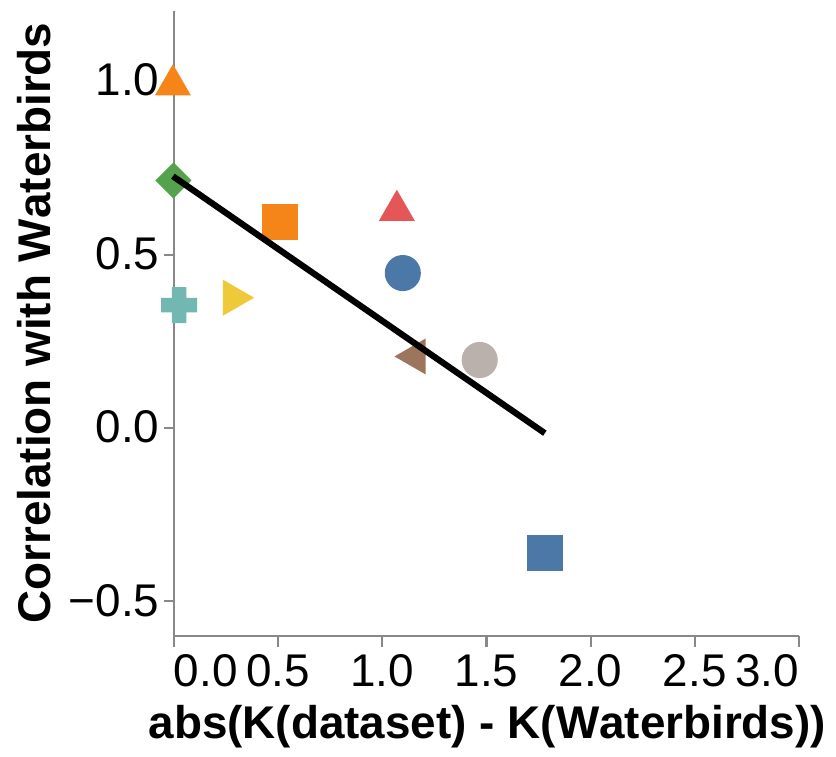}
         \caption{\waterbirds}
     \end{subfigure}
     \hfill
     \begin{subfigure}[b]{0.17\textwidth}
        \centering
        \includegraphics[trim={0 0 0 0},clip,width=\textwidth]{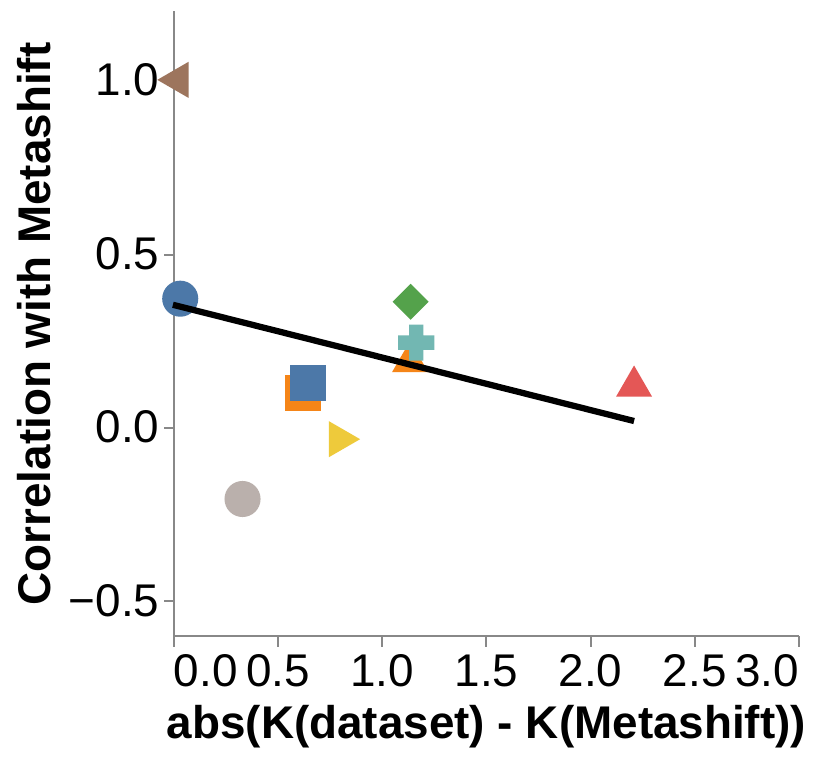}
         \caption{Metashift}
     \end{subfigure}
     \hfill
     \begin{subfigure}[b]{0.17\textwidth}
        \centering
        \includegraphics[trim={0 0 0 0},clip,width=\textwidth]{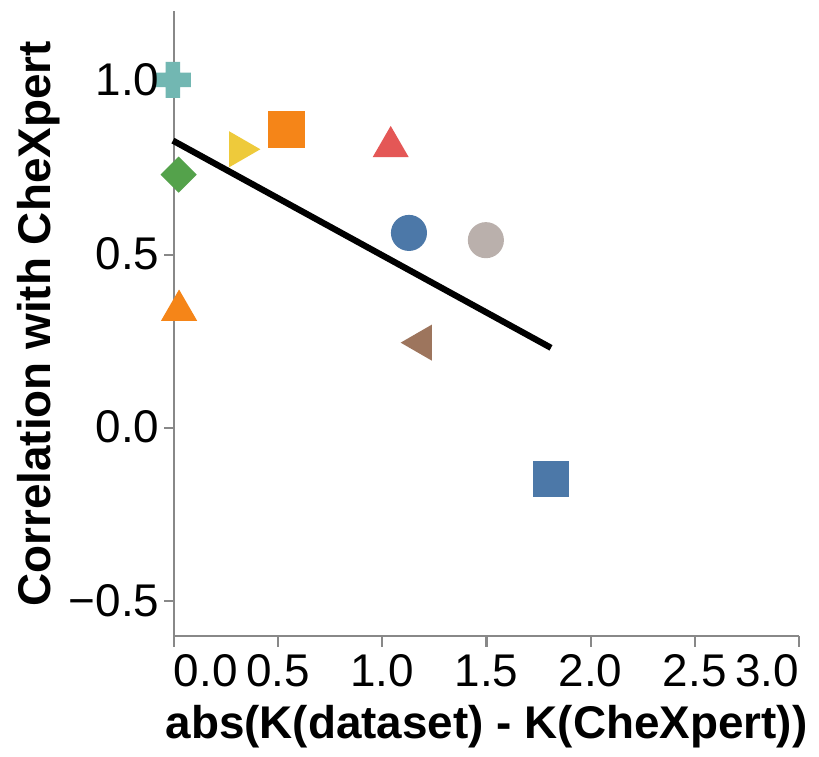}
         \caption{\chexpert}
     \end{subfigure}
     \hfill
     \begin{subfigure}[b]{0.23\textwidth}
        \centering
        \includegraphics[trim={0 0 0 0},clip,width=\textwidth]{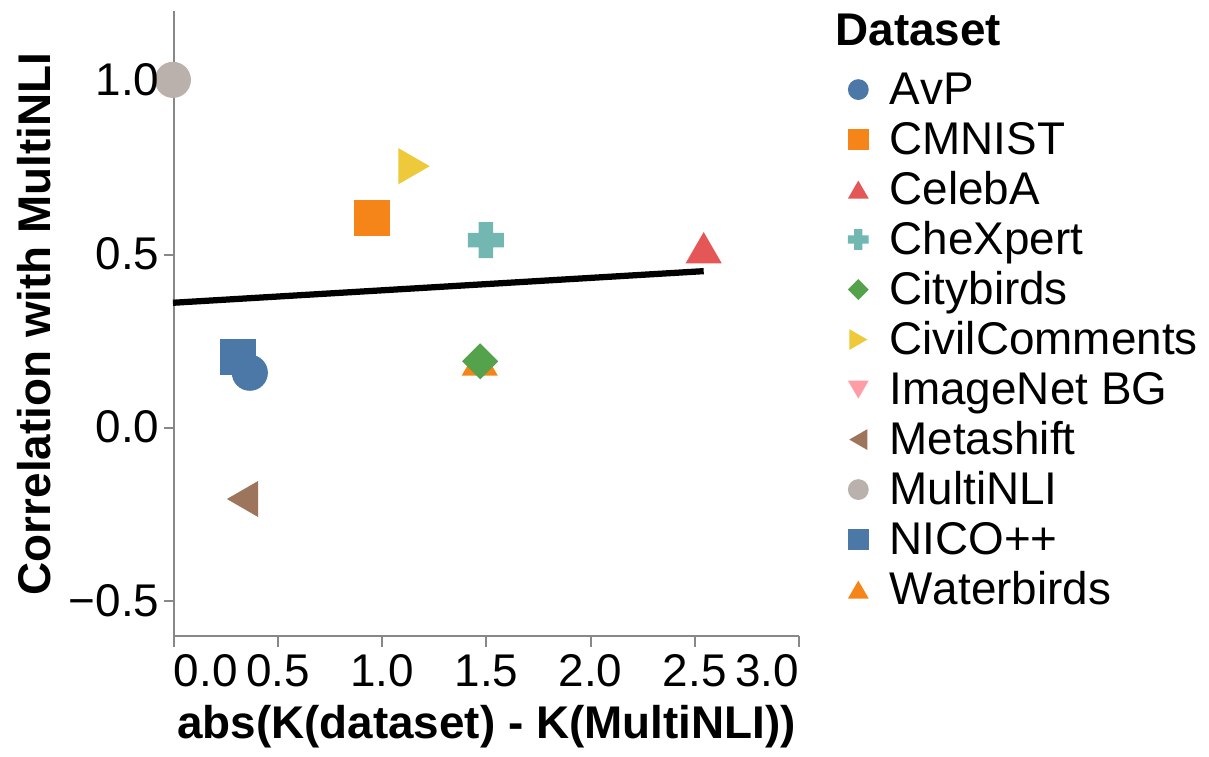}
         \caption{\multinli}
     \end{subfigure}
     
     \begin{subfigure}[b]{0.17\textwidth}
        \centering
        \includegraphics[trim={0 0 0 0},clip,width=\textwidth]{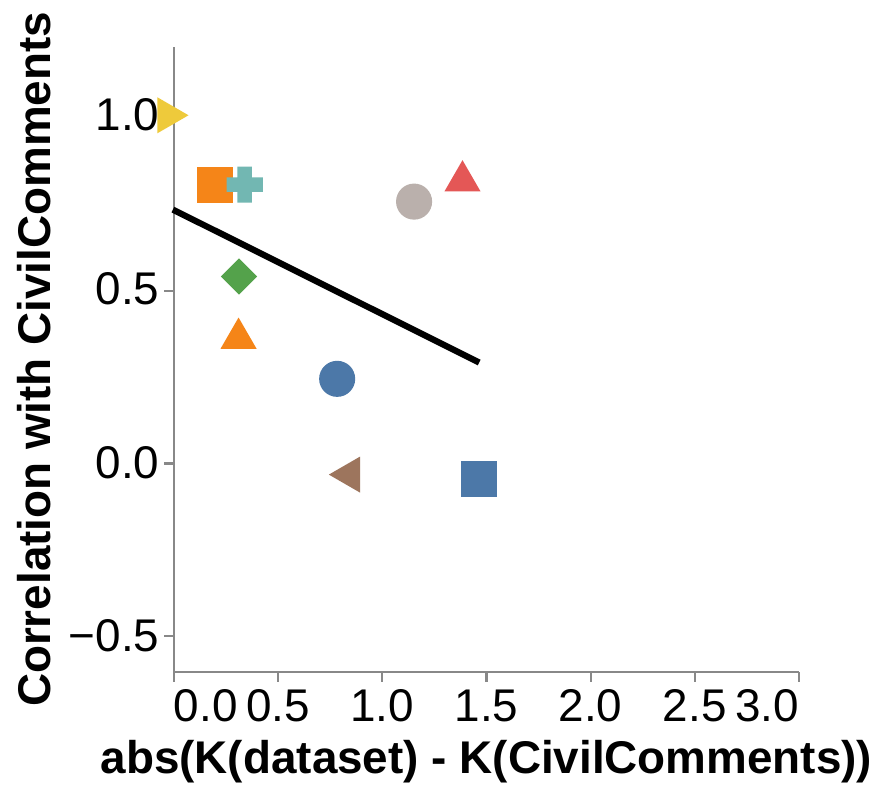}
         \caption{\civilcomments}
     \end{subfigure}
     \hfill
     \begin{subfigure}[b]{0.17\textwidth}
        \centering
        \includegraphics[trim={0 0 0 0},clip,width=\textwidth]{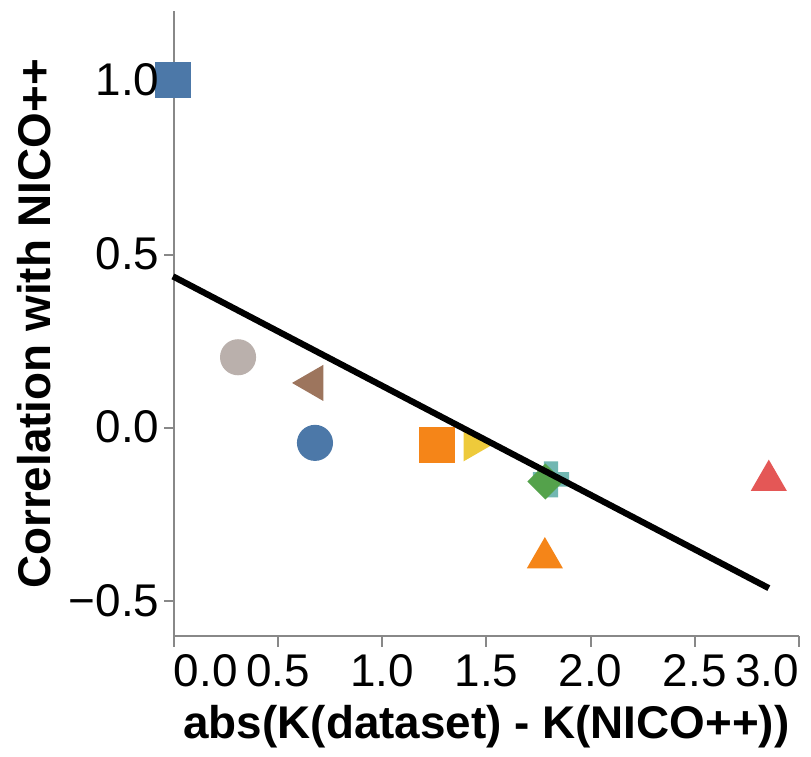}
         \caption{\nicopp}
     \end{subfigure}
     \hfill
     \begin{subfigure}[b]{0.17\textwidth}
        \centering
        \includegraphics[trim={0 0 0 0},clip,width=\textwidth]{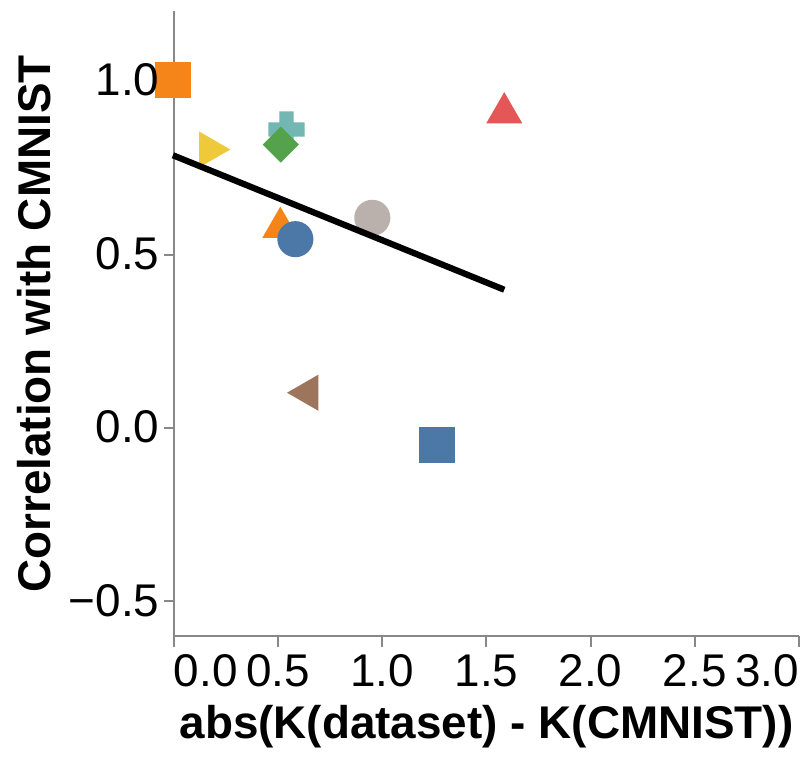}
         \caption{CMNIST}
     \end{subfigure}
     \hfill
     \begin{subfigure}[b]{0.17\textwidth}
        \centering
        \includegraphics[trim={0 0 0 0},clip,width=\textwidth]{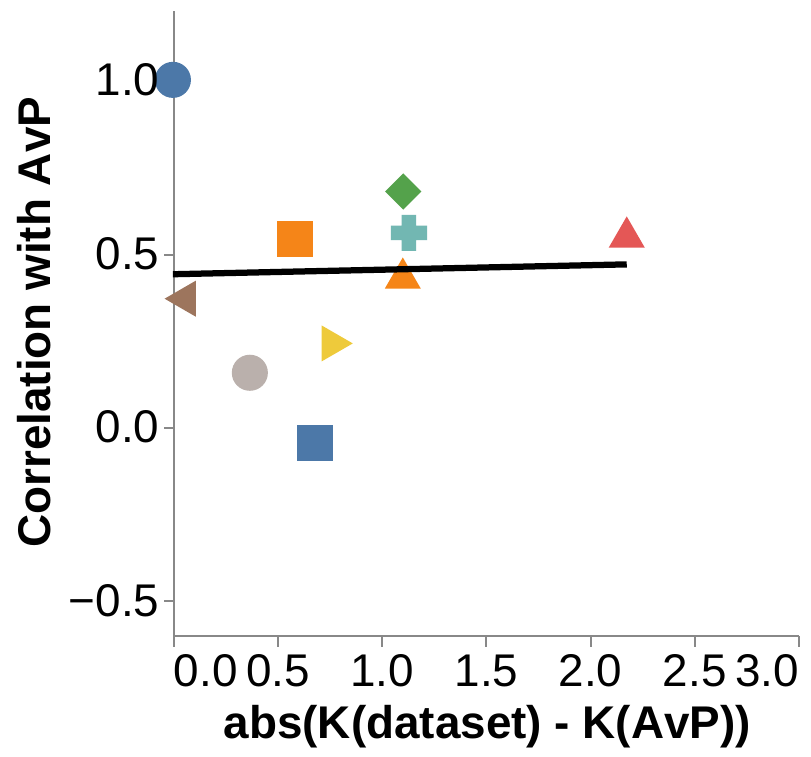}
         \caption{\animalsvsplants}
     \end{subfigure}
     \hfill
     \begin{subfigure}[b]{0.23\textwidth}
        \centering
        \includegraphics[trim={0 0 0 0},clip,width=\textwidth]{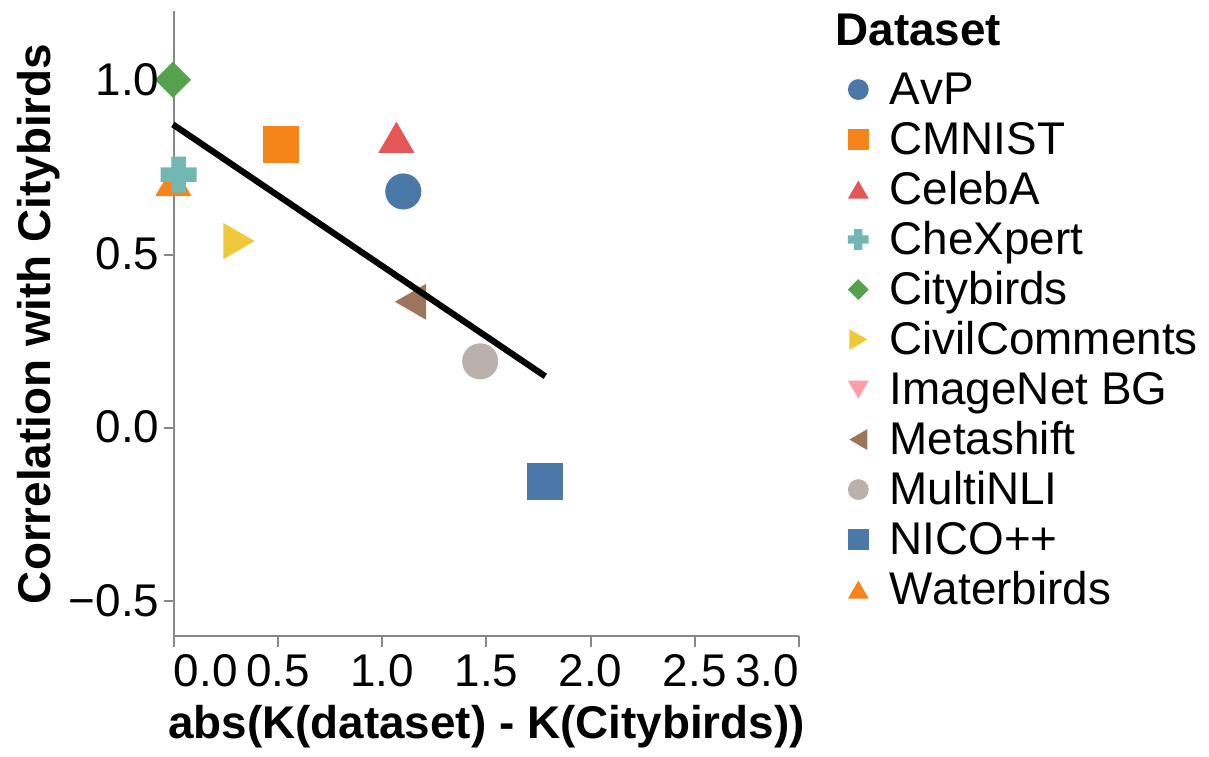}
         \caption{\citybirds}
     \end{subfigure}
    \caption{Benchmark agreement (Pearson's $r$) as a function of difference in task difficulty due to spurious correlation, as measured by Bayes Factor $K$. Each panel shows the agreement in worst-group test accuracies on the named dataset vs. all other datasets. Only benchmarks valid according to ERM group variability and method variability are included. \textbf{Valid benchmarks should agree more strongly with those that exhibit a similar $K$, thus exhibiting a negative correlation.} Black solid line fit with OLS linear regression.}
    \label{fig:bayes-factor-vs-correlation}
\end{figure*}

\subsection{Evaluating Discriminative Power}
\label{sec:des-2-disc-power}

To evaluate \discpower, we test whether benchmarks assign different scores to different methods, by measuring the SD over method worst-group test accuracies for each dataset.
Low SD indicates that all methods perform similarly, so the benchmark cannot discriminate between them, therefore not satisfying \discpower.\footnote{A limitation of our approach is that low \discpower could also be the result of all methods exhibiting similarly poor performance, such that the benchmark is valid but all methods are insufficiently powerful. Given the inclusion of ERM in the set of methods, and assuming \ermfailure is satisfied, we don't imagine this to be of practical concern.}

\Cref{fig:std-erm-std-method-method} shows the SD over methods in \subpopbench. 
\imagenetbg results in tightly-clustered worst-group accuracies, not satisfying \discpower. 
We also observe a strong positive correlation between ERM accuracy variability across groups and worst-group test accuracy variability across methods (\cref{fig:std-erm-std-method-method-both}).
We hypothesize that benchmarks \emph{could} exist that satisfy \ermfailure but not \discpower, but we do not observe them here. 

\subsection{Evaluating Convergent Validity}
\label{sec:des-3-convergent-validity}

We suggest that the defining characteristic of a spurious correlations benchmark is the \emph{task difficulty due to spurious correlation}.
Thus, to test \convergentvalidity we check if two benchmarks with similar task difficulty produce similar results.
Specifically, we measure how inter-benchmark agreement changes as a function of the difference in task difficulty due to spurious correlation.
Datasets that exhibit convergent validity should show a strong correlation, where increasing distance increases disagreement.

\begin{figure*}[tb!]
    \centering
    \begin{subfigure}[t]{0.20\textwidth}
        \centering
        \includegraphics[width=\textwidth]{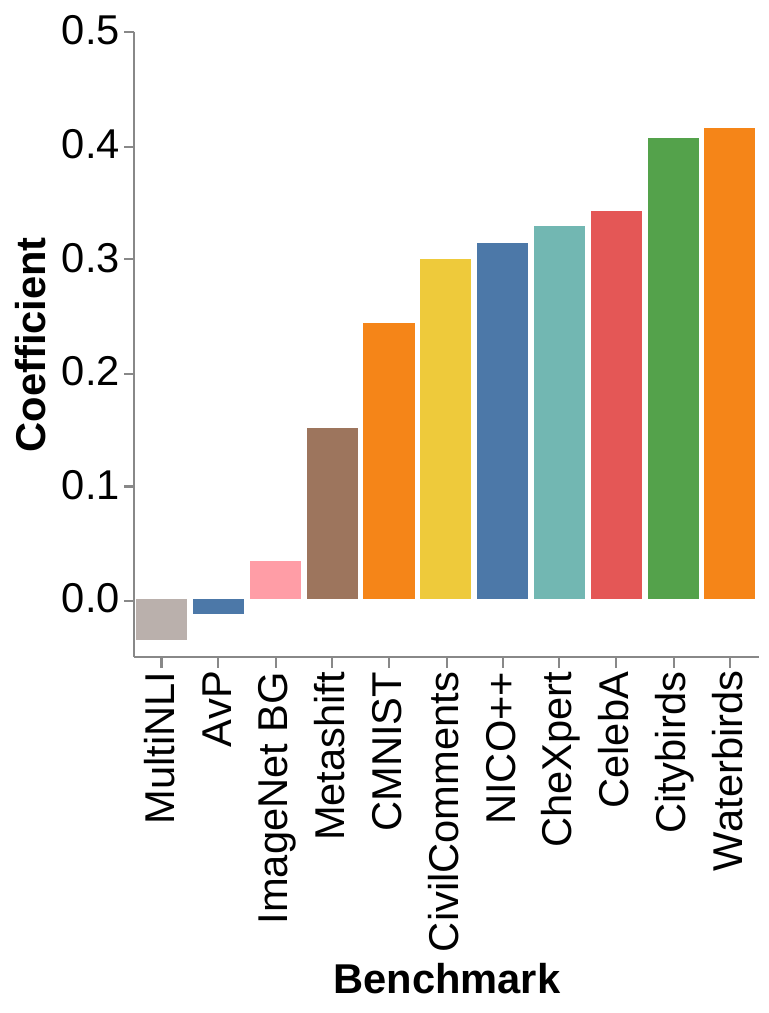}
        \caption{Coefficient}
        \label{fig:dataset-validity-coeff}
    \end{subfigure}
    \hfill
    \begin{subfigure}[t]{0.20\textwidth}
        \centering
        \includegraphics[width=\textwidth]{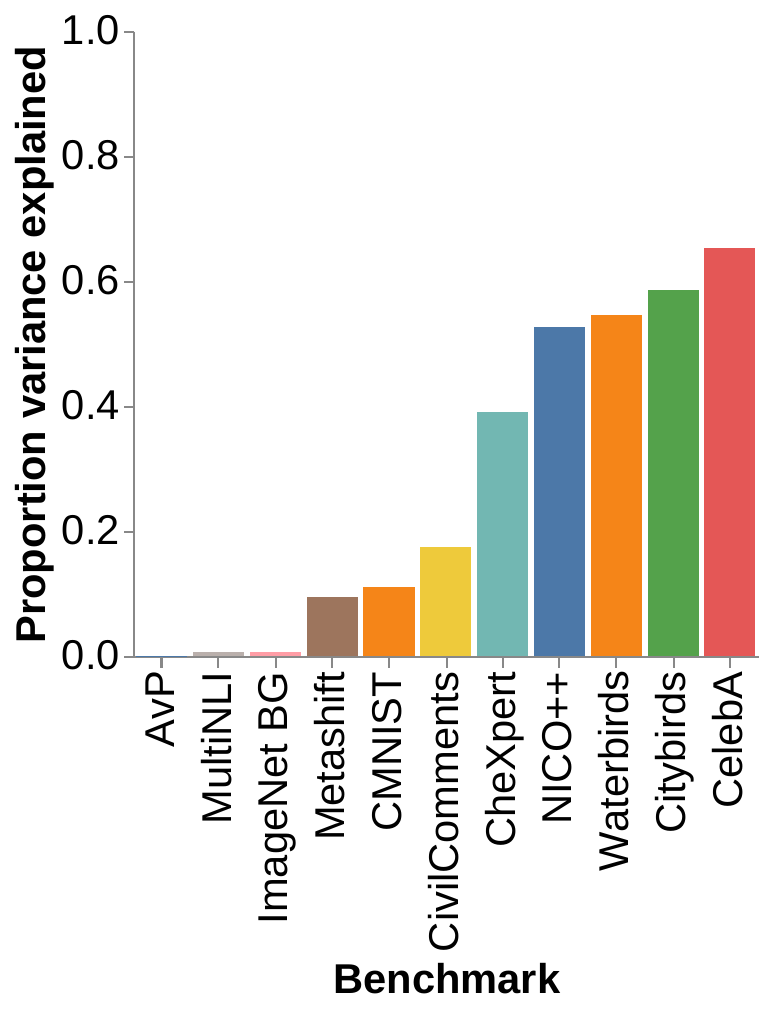}
        \caption{Goodness of fit}
        \label{fig:dataset-validity-rsquared}
    \end{subfigure}
    \hfill
    \begin{subfigure}[t]{0.20\textwidth}
        \centering
        \includegraphics[width=\textwidth]{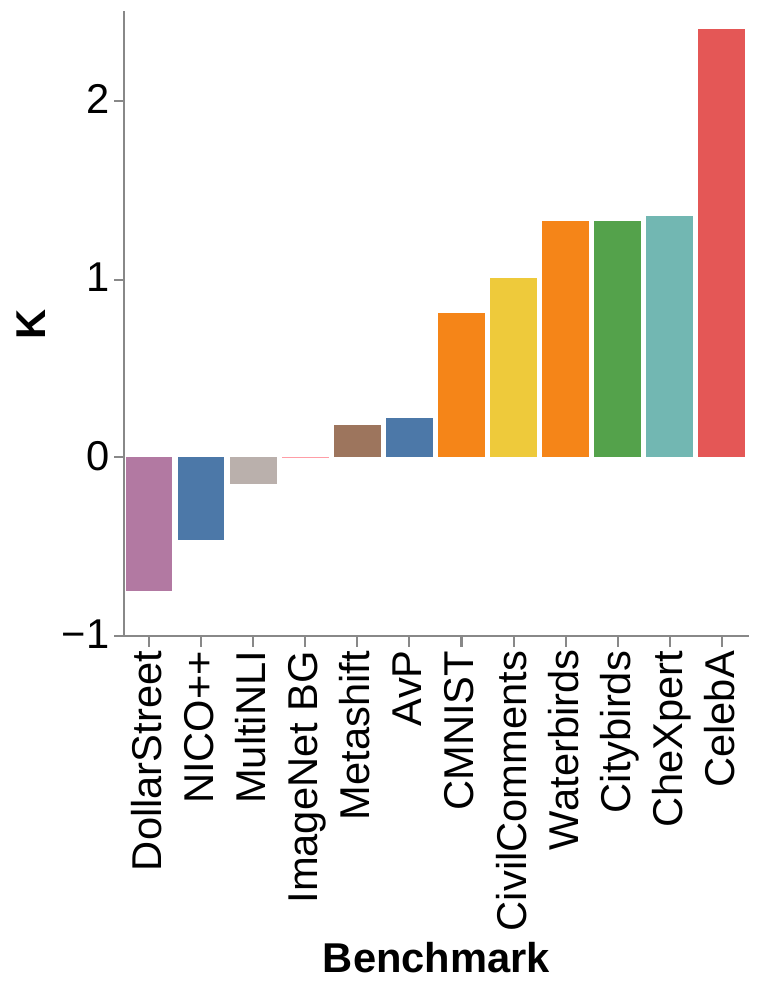}
        \caption{$K$}
        \label{fig:dataset-validity-k}
    \end{subfigure}
    \caption{\textbf{(a)} Benchmark agreement (Pearson's $r$) as a function of by difference in task difficulty due to spurious correlation ($K$). I.e., the (negative) slope of each line in \cref{fig:bayes-factor-vs-correlation}. \textbf{(b)} $R^2$ of each line. \textbf{(c)} Task difficulty due to spurious correlation, $K$. \textbf{Valid benchmarks should most agree with other benchmarks with similar $K$, so a large coefficient indicates a more valid benchmark.}}
    \label{fig:dataset-validity}
\end{figure*}

\newpage

\subsubsection{Introducing \texorpdfstring{$K$}{K}: Task Difficulty due to Spurious Correlation}
\label{sec:measuring-task-difficulty}

Before we can evaluate \convergentvalidity, we must first take a brief detour to understand what benchmarks actually measure.
We argue that three factors should determine how a spurious correlations benchmark behaves: 
\begin{enumerate}
    \itemsep0em 
    \item The strength of the association, i.e. how often targets and attributes co-occur in the data;
    \item The difficulty of learning the correlated attribute, i.e. how easily can a model predict the attribute; and
    \item The difficulty of learning the intended target, i.e. how easily the model can predict the class label.
\end{enumerate}
A common measure of spurious correlation is mutual information (MI) between groups and targets \citep{yang2023change}.
However, MI only accounts for factor 1, and is not sensitive to factors 2 and 3, which are necessarily model-dependent.\footnote{Consider a grayscale-only model trained on Coloured MNIST. While there may be a spurious correlation between target and attribute, the model cannot exploit it \citep{arjovsky2019irm}.}
See \cref{sec:app-mutual-information} for empirical evidence of this problem.
To account for all three factors, we propose using the Bayes Factor as a model-dependent statistic that quantifies the task difficulty due to spurious correlations.
The Bayes Factor evaluates the relative model performance of a model that \emph{can} leverage the spurious correlation to solve the task, and a model that is penalized for doing so. 

We use the Bayes Factor to compare how well the original and penalized models explain the data, defining the task difficulty due to spurious correlation of a benchmark as
\begin{align*}
    K = \log \frac{P(Y^{\mathrm{WG}}_{\mathrm{test}}|X^{\mathrm{WG}}_{\mathrm{test}},M_{RW})}{P(Y^{\mathrm{WG}}_{\mathrm{test}}|X^{\mathrm{WG}}_{\mathrm{test}},M_{ERM})} \ ,
\end{align*}
where the numerator is the likelihood of the worst test group $(Y^{\mathrm{WG}}_{\mathrm{test}}, X^{\mathrm{WG}}_{\mathrm{test}})$, according to the model penalized for using the spurious correlation, $M_{RW}$, and the denominator is according to the model that uses the spurious correlation, $M_{ERM}$.
$M_{ERM}$ is the benchmark's base model trained using ERM, whereas $M_{RW}$ is the same base model trained with a reweighted loss function, where each sample's weight is inversely proportional to its group size.
We choose reweighting for its simplicity and minimal assumptions.
In \cref{sec:app-k-implementation} we present almost identical results using an implementation of $K$ with GroupDRO, additionally finding $K$ is highly robust to hyperparameter tuning for both $M_{ERM}$ and $M_{RW}$.
We use ResNet-50 \citep{he2016resnet} as a base model for vision benchmarks, and BERT for language \citep{devlin2019bert}.

One can interpret $K$ as measuring how much better the reweighted model explains the test set, versus the ERM model.
High $K$ indicates that the task is made more difficult by the spurious correlation, and a low $K$ indicates that the spurious correlation is not a dominant factor. 

\begin{table*}[bt!]
\caption{Measures of validity, \ermfailure, \discpower, and \convergentvalidity, alongside best performing method and resulting worst-group test accuracies, for nine benchmarks from \subpopbench and our two additions. \textbf{Only certain benchmarks satisfy all three desiderata.}}
\label{table:best-method-per-dataset}
\footnotesize
\centering
\begin{tabular}{lrrrrlr}
\toprule
Benchmark & K & ERM Fail. & Disc. Power & Conv. Validity& Method & WG test acc. \\
\midrule
MultiNLI & -0.15 & {\cellcolor[HTML]{F7A98B}} \color[HTML]{000000} 10.81 & {\cellcolor[HTML]{EFCFBF}} \color[HTML]{000000} 10.89 & {\cellcolor[HTML]{B40426}} \color[HTML]{F1F1F1} -0.04 & GroupDRO & 75.58 \\
\animalsvsplants & 0.22 & {\cellcolor[HTML]{E7745B}} \color[HTML]{F1F1F1} 6.47 & {\cellcolor[HTML]{C12B30}} \color[HTML]{F1F1F1} 2.96 & {\cellcolor[HTML]{C53334}} \color[HTML]{F1F1F1} -0.01 & GroupDRO & 91.75 \\
\imagenetbg & -0.01 & {\cellcolor[HTML]{B40426}} \color[HTML]{F1F1F1} 0.65 & {\cellcolor[HTML]{B40426}} \color[HTML]{F1F1F1} 2.16 & {\cellcolor[HTML]{E46E56}} \color[HTML]{F1F1F1} 0.03 & CRT & 78.22 \\
Metashift & 0.18 & {\cellcolor[HTML]{F59F80}} \color[HTML]{000000} 9.84 & {\cellcolor[HTML]{C12B30}} \color[HTML]{F1F1F1} 2.95 & {\cellcolor[HTML]{EFCEBD}} \color[HTML]{000000} 0.15 & ReSample & 80.00 \\
CMNIST & 0.80 & {\cellcolor[HTML]{3B4CC0}} \color[HTML]{F1F1F1} 35.68 & {\cellcolor[HTML]{4257C9}} \color[HTML]{F1F1F1} 22.45 & {\cellcolor[HTML]{BAD0F8}} \color[HTML]{000000} 0.24 & LISA & 71.09 \\
CivilComments & 1.00 & {\cellcolor[HTML]{D3DBE7}} \color[HTML]{000000} 19.40 & {\cellcolor[HTML]{F7A688}} \color[HTML]{000000} 8.06 & {\cellcolor[HTML]{8FB1FE}} \color[HTML]{000000} 0.30 & GroupDRO & 72.53 \\
NICO++ & -0.47 & {\cellcolor[HTML]{F7AC8E}} \color[HTML]{000000} 11.14 & {\cellcolor[HTML]{F29274}} \color[HTML]{F1F1F1} 7.00 & {\cellcolor[HTML]{84A7FC}} \color[HTML]{F1F1F1} 0.31 & Focal & 37.78 \\
CheXpert & 1.35 & {\cellcolor[HTML]{5F7FE8}} \color[HTML]{F1F1F1} 31.52 & {\cellcolor[HTML]{3B4CC0}} \color[HTML]{F1F1F1} 22.94 & {\cellcolor[HTML]{779AF7}} \color[HTML]{F1F1F1} 0.33 & CBLoss & 75.08 \\
CelebA & 2.39 & {\cellcolor[HTML]{AEC9FC}} \color[HTML]{000000} 23.61 & {\cellcolor[HTML]{CBD8EE}} \color[HTML]{000000} 13.90 & {\cellcolor[HTML]{6E90F2}} \color[HTML]{F1F1F1} 0.34 & DFR & 87.78 \\
Citybirds & 1.32 & {\cellcolor[HTML]{EDD1C2}} \color[HTML]{000000} 15.57 & {\cellcolor[HTML]{EED0C0}} \color[HTML]{000000} 11.01 & {\cellcolor[HTML]{3F53C6}} \color[HTML]{F1F1F1} 0.41 & ReWeight & 90.50 \\
Waterbirds & 1.32 & {\cellcolor[HTML]{F7BA9F}} \color[HTML]{000000} 12.65 & {\cellcolor[HTML]{E36C55}} \color[HTML]{F1F1F1} 5.31 & {\cellcolor[HTML]{3B4CC0}} \color[HTML]{F1F1F1} 0.41 & LISA & 86.98 \\
\bottomrule
\end{tabular}
\vspace{-8pt}
\end{table*}

\subsubsection{Sanity Checking \texorpdfstring{$K$}{K}}

To ensure that $K$ functions correctly, we test it using artificially manipulated datasets.
We produce versions of \celeba and \waterbirds with increasingly correlated attributes and targets (``confounder strength''), and versions of \waterbirds with various amounts of background noise, foreground noise and attribute noise (see \cref{sec:app-validating-k}).
\Cref{fig:bayes-factor} shows that, as intended, $K$ increases with confounder strength on both \waterbirds and \celeba (\cref{fig:bayes-factor-celeba-confounder,fig:bayes-factor-wb-confounder}).
On \waterbirds, $K$ decreases as background noise increases (reducing the utility of the spurious correlation), as long as a correlation exists (\cref{fig:bayes-factor-wb-bg-noise,fig:bayes-factor-wb-bg-noise-no-corr}).
Conversely, $K$ increases as foreground noise increases (increasing the utility of the spurious correlation; \cref{fig:bayes-factor-wb-attribute-noise}). 
We also see that \citybirds{} has a $K$ equal to that of \waterbirds{}, and \animalsvsplants has low $K$ reflecting the limited utility of the spuriously correlated geographic information (\cref{fig:dataset-validity-k}). 
Our experiments support $K$ as a measure of task difficulty due to spurious correlation, though we note that $K$ depends on both the availability and quality of attribute annotations (\cref{fig:bayes-factor-wb-attribute-noise}).\footnote{We consider 100\% attribute noise equivalent to attribute information not being available.}

\subsubsection{Measuring Convergent Validity with \texorpdfstring{$K$}{K}}

\Cref{fig:dataset-validity-k} shows the value of $K$ for our sample of benchmark datasets.
Certain benchmarks, e.g. \celeba, have very high $K$, indicating that task difficulty is dominated by the spurious correlations in the data.
Other datasets, such as \dollarstreet and \nicopp, exhibit low $K$, suggesting spurious correlation is not a principal factor in task difficulty.

Recall in \cref{sec:des-3-convergent-validity} we said that benchmarks exhibiting \convergentvalidity must agree with other benchmarks that measure the same thing. 
More precisely, if we use $K$ to describe what a benchmark measures, valid benchmarks should agree most strongly with other datasets with a similar task $K$.
In other words, between-benchmark disagreement should increase as two datasets have more different values of $K$. 
\Cref{fig:bayes-factor-vs-correlation} shows the the agreement (Pearson's $r$) in worst-group test accuracies achieved over methods between pairs of benchmarks, as a function of the distance in $K$ between the benchmarks.
A strong negative slope indicates that the more benchmarks differ, the more they disagree, whereas a horizontal slope indicates that agreement is not a function of $K$ difference, i.e. not satisfying convergent validity.
\Cref{fig:dataset-validity-coeff} shows the (negative) slope of the agreement function for each benchmark.
\multinli, \animalsvsplants and \imagenetbg have low \convergentvalidity.

\textbf{Summary.} \imagenetbg satisfies none of our three desiderata, whereas \multinli and \animalsvsplants satisfy \ermfailure and \discpower but not \convergentvalidity. See \cref{table:best-method-per-dataset}.

\section{Not All Methods are Robust}
\label{sec:method-robustness}

\begin{figure*}[t]
    \centering
    \begin{subfigure}[t]{0.25\textwidth}
        \centering
        \includegraphics[width=\textwidth,valign=t]{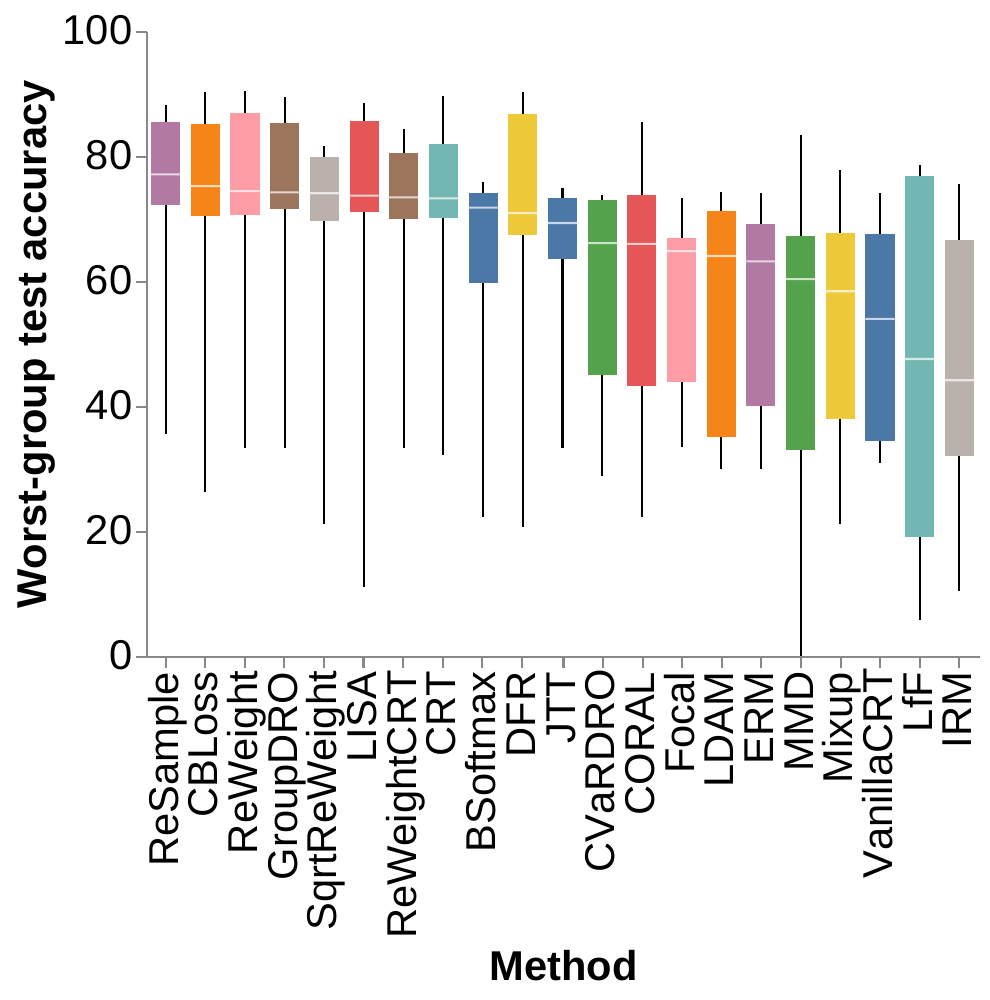}
        \vspace{1pt}
        \caption{WG test accuracies}
        \label{fig:method-sensivity-box}
    \end{subfigure}
    \hfill
    \begin{subfigure}[t]{0.35\textwidth}
        \centering
        \includegraphics[width=\textwidth,valign=t]{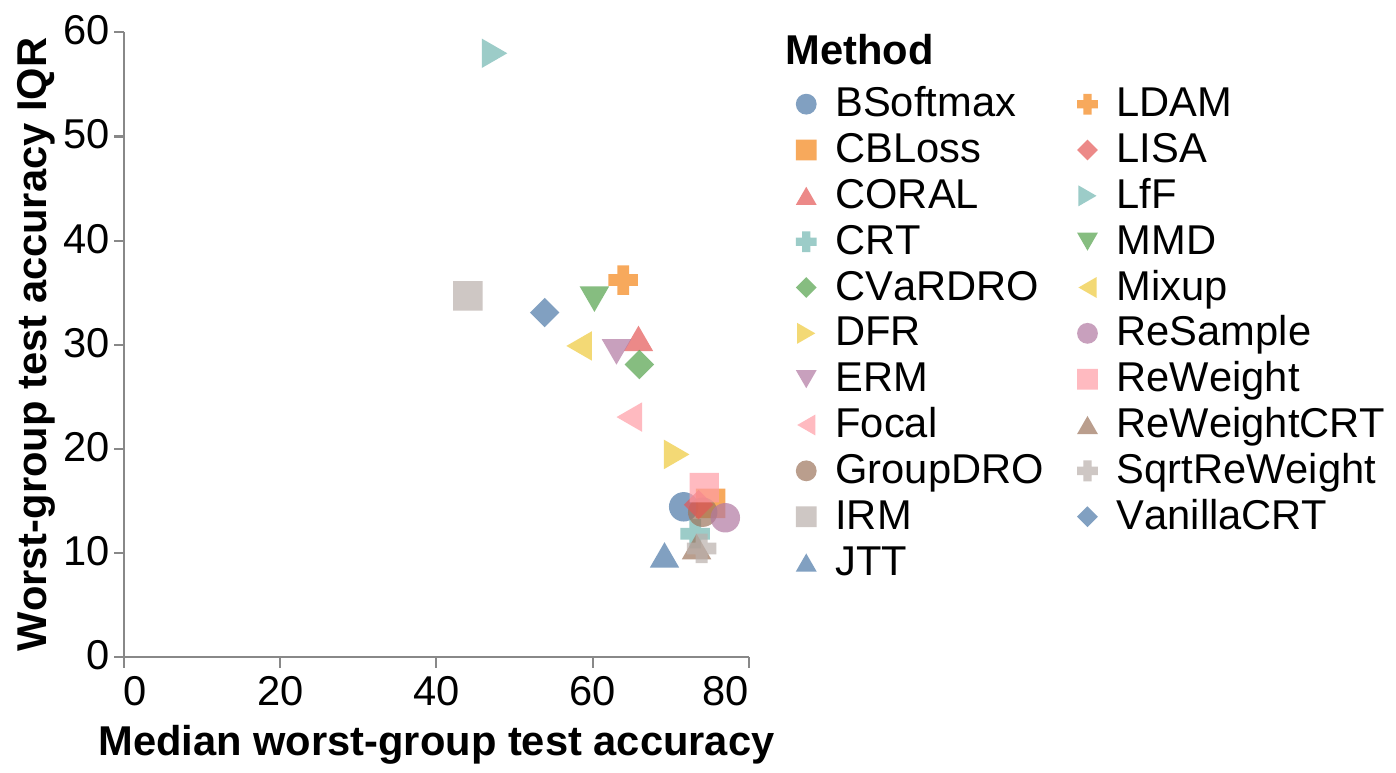}
        \vspace{28pt}
        \caption{Performance and variability}
        \label{fig:method-sensivity-median-iqr}
    \end{subfigure}
    \hfill
    \begin{subfigure}[t]{0.25\textwidth}
        \centering
        \includegraphics[width=\textwidth,valign=t]{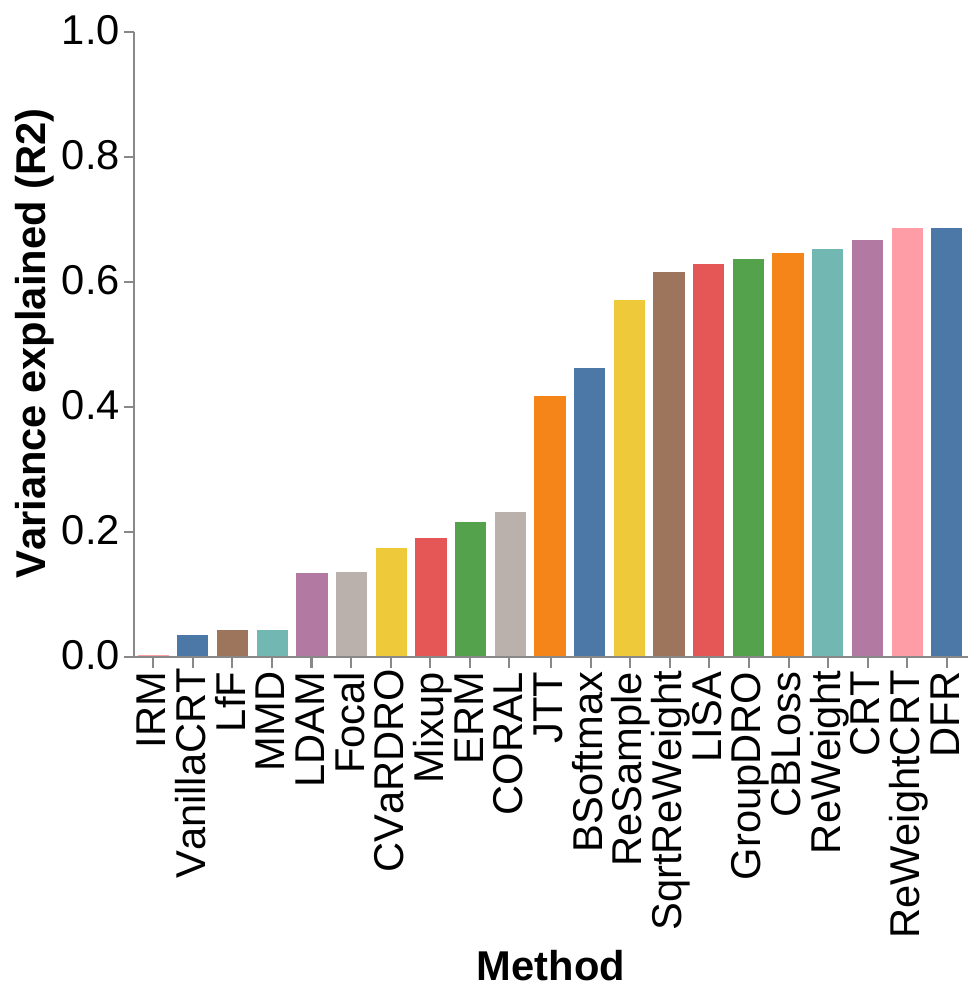}
        \caption{Variance explained}
        \label{fig:method-sensivity-rsquared}
    \end{subfigure}
    \caption{\textbf{(a)} Box plot of worst-group test accuracies for each method over all datasets. White line  median; box IQR; whiskers range. \textbf{(b)} Median and IQR of worst-group test accuracies for each method. \textbf{(c)} Proportion of worst-group test accuracy variance explained ($R^2$) by $K$ for each method over all datasets. \textbf{Only certain methods consistently yield high worst-group accuracies. Of stable methods, benchmark $K$ explains a large proportion of variance.}}
    \label{fig:method-senstivity}
\end{figure*}

The function of a benchmark is to evaluate method efficacy to support reasoning about which method to use.
Having discarded certain benchmarks as invalid measures, we now ask whether some methods are more robust than others. 
As a step towards making practical recommendations in real-world contexts, we ask whether certain methods are more robust to benchmarks with different $K$.

\Cref{fig:method-sensivity-box} shows the distribution of worst-group test accuracies over benchmarks for each method, while \cref{fig:method-sensivity-median-iqr} shows the median worst-group test accuracy and the inter-quartile range (IQR).\footnote{We use median and IQR as many methods exhibit non-normal performance distributions over benchmarks.}
Methods in the lower-right corner of \cref{fig:method-sensivity-median-iqr}, such as CRT, ReWeight, ReSample and GroupDRO, exhibit both high performance and low variability over benchmarks.

Next, we ask to what extent method variability is a function of the benchmark's task difficulty due to spurious correlation.
\Cref{fig:method-sensivity-rsquared} shows how much a benchmark's $K$ explains the variance in worst-group test accuracies (see \cref{sec:app-method-sensitivity}).
Many of the methods with low IQRs have a large proportion of their variance explained by the benchmark's $K$, suggesting that while variability may be reduced, they remain sensitive to benchmark specifics.

\textbf{Summary.} While practitioners must consider the specific nature of their dataset, certain methods, e.g.\ CRT, ReWeight, ReSample and GroupDRO, achieve consistently high worst-group test accuracies.

\section{Practical Recommendations}
\label{sec:practical-recommendations}

When competing benchmarks suggest different methods, practitioners are left with a crucial question: what should I use for \emph{my} specific dataset?
Building upon our investigation of benchmark validity, we take a first step towards bridging the gap between real-world problems and common benchmarks.

A typical strategy is to take the method with the best performance averaged over all benchmarks.
We evaluate two improvements over this conventional practice. 
First, we suggest only averaging over benchmarks deemed valid according to our three desiderata.
Second, we suggest choosing the most similar benchmark to our given problem.
Concretely, given some arbitrary dataset (which we refer as a ``test dataset''\footnote{We describe these as test datasets, rather than benchmarks, to emphasize we are evaluating our approach \emph{as if we were a practitioner}, faced with some new test dataset.}) we recommend that practitioners first calculate its $K$, and select the benchmark with the closest $K$.
We hypothesize that the best performing methods on the closest benchmark will be most appropriate for the test dataset.

We test our approach using a Leave-One-Out analysis, where for each of eight test datasets we select the next closest benchmark, and evaluate the closest benchmark's best performing method on our test dataset (``Closest Benchmark''). 
We compare worst-group test accuracies for this method against the best method according to the all-benchmark average (``All Benchmarks''), and the valid-benchmark average (``Valid Benchmarks'').
We perform an additional evaluation using \dollarstreet, a dataset of geographically-diverse household images.
Previous work has identified significantly worse performance of contemporary vision models for non-Western regions \citep{devries2019does, richards2023progress}.
We extend \subpopbench to include a 42-class object classification task using \dollarstreet, where group information is geographic region.
See \cref{sec:app-benchmark-datasets} for further details.

\Cref{table:predicted-method} shows the results of our two approaches.
Averaging over benchmarks results in a single ``best'' method.
Over all benchmarks, this method is GroupDRO; over valid benchmarks only this is ReSample. 
For six of nine test datasets, ReSample outperforms GroupDRO, confirming that filtering benchmarks before averaging is helpful. 
For five of eight test datasets,\footnote{We exclude \civilcomments as it lacks appropriate comparison benchmarks. 
Recall that we found \multinli to be invalid, and that $K$ is model-specific, such one can't compare across a ResNet-based and a BERT-based $K$.\label{foot:civilcomments}} selecting a method according to the closest benchmark would produce superior worst-group test accuracies, supporting our hypothesis that selecting based on similarity improves performance.

\begin{table*}[bt!]
\caption{Results of selecting a method according to worst-group test accuracy averaged over all benchmarks ("All Benchmarks"), averaged over valid benchmarks ("Valid Benchmarks"), and on the closest valid benchmark according to $K$ (``Closest Benchmark''). \textbf{Using the closest benchmark (col.~8) results in the best performance for 5 out of 8 test datasets}\footref{foot:civilcomments}. Alternatively, averaging over valid benchmarks (col.~5) instead of all benchmarks (col.~3) improves performance on 6 out of 9 test datasets. \textbf{On \dollarstreet, using the closest benchmark improves on standard practice.}} 
\label{table:predicted-method}
\footnotesize
\centering
\begin{tabular}{llrlrllr}
\toprule
 & \multicolumn{2}{c}{All Benchmarks} & \multicolumn{2}{c}{Valid Benchmarks} & \multicolumn{3}{c}{Closest Benchmark} \\
\cmidrule(lr){2-3} \cmidrule(lr){4-5} \cmidrule(lr){6-8}
Test Dataset & Method & Acc. & Method & Acc. & Closest benchmark & Method & Acc. \\
\midrule
CelebA & GroupDRO & 87.22 & ReSample & 85.37 & CheXpert & CBLoss & \bfseries 87.41 \\
CheXpert & GroupDRO & 71.82 & ReSample & \bfseries 74.21 & Citybirds & ReWeight & 73.95 \\
Citybirds & GroupDRO & \bfseries 89.46 & ReSample & 88.16 & Waterbirds & LISA & 88.52 \\
CivilComments & GroupDRO & \bfseries 72.53 & ReSample & 72.50 & - &  - & -  \\
CMNIST & GroupDRO & 70.83 & ReSample & 70.99 & Waterbirds & LISA & \bfseries 71.09 \\
Metashift & GroupDRO & 75.90 & ReSample & \bfseries 80.00 & CMNIST & LISA & 72.31 \\
NICO++ & GroupDRO & 33.33 & ReSample & \bfseries 35.56 & Metashift & ReSample & \bfseries 35.56 \\
Waterbirds & GroupDRO & 84.79 & ReSample & 85.57 & Citybirds & ReWeight & \bfseries 86.68 \\
\midrule
DollarStreet & GroupDRO & 58.65 & ReSample & 58.81 & NICO++ & Focal & \bfseries 59.25 \\
\bottomrule
\end{tabular}
\vspace{-12pt}
\end{table*}

On \dollarstreet, selecting a method by averaging over valid benchmarks improves over indiscriminate averaging, and selecting a method using the closest benchmark improves performance further.
We highlight that in \cref{fig:dataset-validity-k} we see a low $K$ for \dollarstreet, which suggests that it is \emph{not} a dataset dominated by spurious correlations, and other factors may contribute \citep{gustafson2023exploring}. 

\textbf{Summary.} Practitioners should only average over valid benchmarks when determining which method is best. Choosing methods according to the closest benchmark may improve performance.

\section{Discussion}
\label{sec:discussion}
Our work joins a wider discussion evaluation practice validity, such as that of 
\citet{jacobs2021measurement} who argue that machine learning researchers can ``collapse the distinctions between constructs and their operationalizations'' (p.~384).
\citet{blodgett2021stereotyping} find that many NLP fairness benchmarks are not meaningful measurement tools, while \citet{subramonian2023it} consider how differing task conceptualizations lead to benchmark disagreement.
\citet{denton2021imagenet} examine the implicit assumptions behind ImageNet, in particular the way its creators operationalize a particular conception of vision. 
Benchmarks considered here may encode notions of spuriousness whose appropriateness is context-sensitive.

One possible challenge to validity is that benchmarks are often adapted and repurposed \citep{koch2021reduced}.
For example, our results show that \multinli does not satisfy our three desiderata for spurious correlations benchmarks. 
This is to some extent expected: our results should not be interpreted as a critique of \multinli, but of its adoption for an unintended purpose \citep{williams2018multinli}. 
Developing meaningful benchmarks is a challenging feat, and one unlikely to be accomplished by reaching for what is most convenient.

When simplifying real-world problems into benchmarks, we sometimes lose sight of our original intent \citep{selbst2019abstraction}. 
Our results call for consideration of whether solving a benchmark corresponds to solving a real problem.
For example, we observe that \dollarstreet has a low task difficulty due to spurious correlation (\cref{fig:dataset-validity-k}).
Thus, if our motive were to reduce computer vision's Western bias \citep{richards2023progress}, one could ask whether developing methods optimized for spurious correlations is a helpful endeavor.

Beyond validity, we might also ask about benchmark \emph{acceptability}.
One benchmark, \celeba \citep{liu2015celeba}, involves an association between hair color and a binary ``gender'' attribute (according to an external annotator), reinforcing views about gender that could be harmful to non-binary and gender nonconforming people \citep{os2018misgendering, denton2020image}.
It is our firm view that \celeba should not be used for benchmarking.

One limitation of our work is that both the \ermfailure and \discpower desiderata depend on well-defined group attributes.
Similarly, while \convergentvalidity can in principal be applied to any benchmark, our use of $K$ also requires group information for the reweighted model. 
Exploration of spurious correlations benchmarks without attributes remains an exciting area for research.

The need for attribute information is also suggestive of a more subtle limitation. 
It is important to draw a distinction between satisfying a desideratum itself, and passing our test as currently implemented. 
For example, our test for \discpower relies on standard deviation, which can be influenced by the presence of outliers.
Whichever statistic used, reasoning about validity necessitates carefully assessing multiple streams of evidence, such as considering \discpower in combination with other desiderata.

Benchmarks form an essential component of the way machine learning evaluates methods and draws conclusions.
In the domain of spurious correlations, our analysis suggests that not all of our benchmarks yield equally meaningful results. 
We hope to have shown that benchmark choice matters: it leads to different conclusions; different recommendations; and ultimately better or worse deployed models.

\printbibliography

\newpage
\appendix

\section{Meta-analysis of Published Data}
\label{sec:app-published-data}

\begin{figure*}[h!]
    \centering
    \includegraphics[trim={0 0 0 0},clip,width=0.28\textwidth]{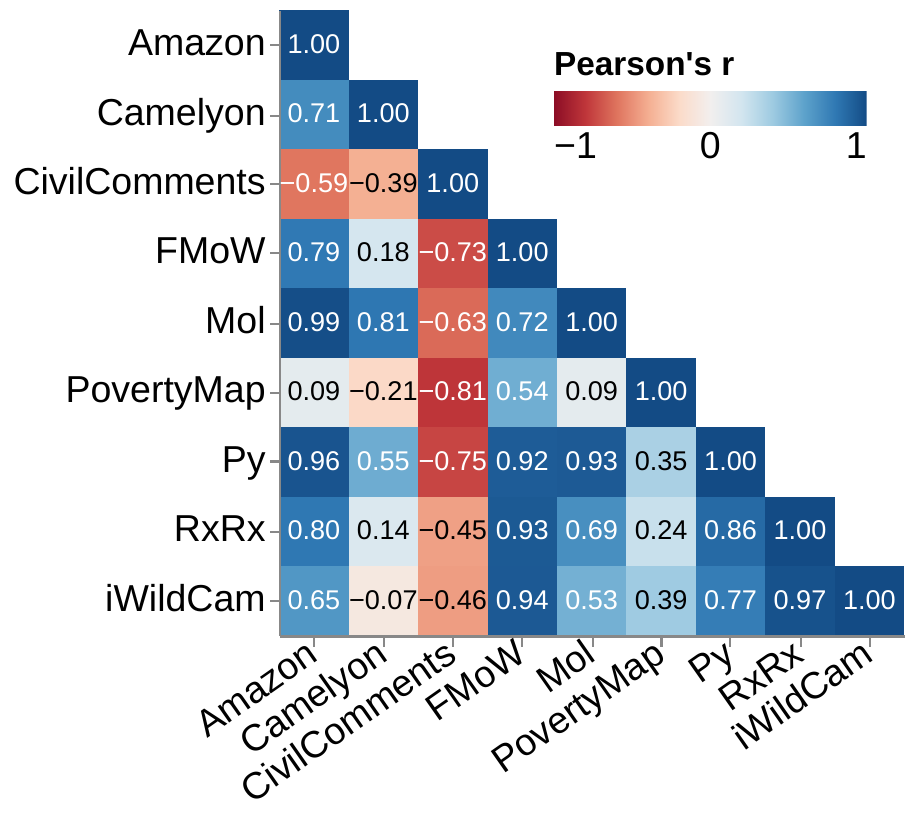}
    \vspace{-8pt}
    \caption{Correlation (Pearson's $r$) between worst-group accuracies across benchmarks in WILDS \citet{koh2021wilds}.}
    \label{fig:published-results-wilds}
\end{figure*}

In \cref{fig:published-results} we analyze previously-published results from the 
\subpopbench benchmarking library \citep{yang2023change}.
We report similar findings analyzing the previously published results from the WILDS benchmarking library \citep{koh2021wilds} in \cref{fig:published-results-wilds}.
We take published worst-group test accuracies, as reported in these two papers, and compute the correlation (Pearson's $r$) between benchmarks. 
For SubpopBench, we use the ``Worst Acc.'' columns from tables reported in \citet[Appendix~E.1]{yang2023change}.
For WILDS, we use the results reported in \citet[table~2]{koh2021wilds}.

\section{Mutual Information}
\label{sec:app-mutual-information}
\begin{figure*}[h!]
    \captionsetup[subfigure]{justification=centering}
    \centering
    \begin{subfigure}[t]{0.16\textwidth}
        \centering
        \includegraphics[trim={0 0 0 0},clip,width=\textwidth]{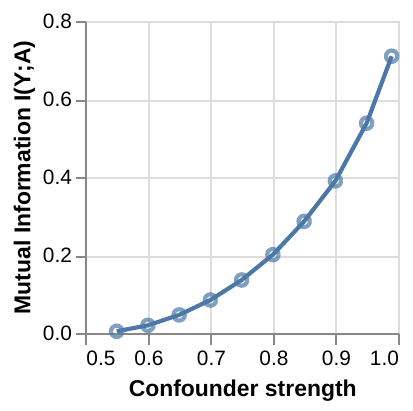}
        \caption{WB confounder strength}
    \end{subfigure}
    \hspace{20pt}
    \begin{subfigure}[t]{0.16\textwidth}
        \centering
        \includegraphics[trim={0 0 0 0},clip,width=\textwidth]{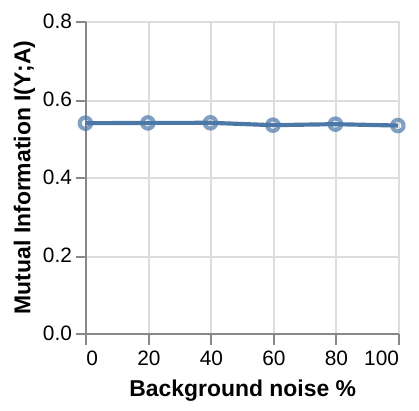}
        \caption{WB\\background noise}
    \end{subfigure}
    \hspace{20pt}
    \hspace{20pt}
    \begin{subfigure}[t]{0.16\textwidth}
        \centering
        \includegraphics[trim={0 0 0 0},clip,width=\textwidth]{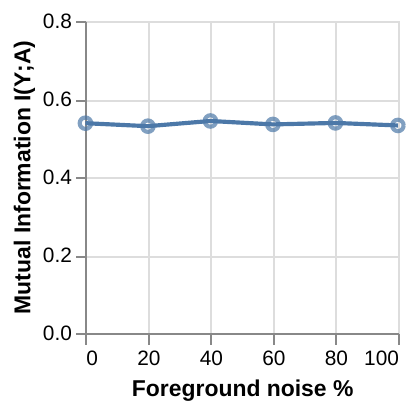}
        \caption{WB foreground noise}
    \end{subfigure}
    \caption{Mutual information $I(Y;A)$ for datasets with various synthetic modifications. By definition, increasing label-attribute correlation increases mutual information (a, b), but cannot account for the effect of background noise (c, d), foreground noise (e) or attribute noise (f). \textbf{Model-independent, data-only metrics are unable to capture the key factors driving task difficulty due to spurious correlation.}}
    \label{fig:mutual-information}
\end{figure*}

Mutual Information (MI) between attributes and target labels is a common metric for evaluating the strength of a spurious correlation. 
Given a group-annotated dataset $(X,Y,A)$ where $X$ is the input data, $Y$ are the target labels and $A$ are group annotations, mutual information $I(Y;A)$ is defined as
\begin{equation}
\label{eq:MI}
I(Y;A) = \sum_{y,a} P(y,a) \log \frac{P(y,a)}{P(y)P(a)} .
\end{equation}
However, mutual information will fail to account for the relative difficulty of learning to predict the target, or vice-versa learning to use the spurious correlation. 
As an illustrative example, consider a version of Waterbirds \citep{sagawa2020gdro} where the background is 100\% noise.
Whatever the mutual information between attributes and labels, this would not be a practically important spurious correlation. 
Conversely, in a dataset where the image of the bird is 100\% noise, the task would be so difficult as to increase the reliance on the spurious correlation. 
We demonstrate this, showing the effect of applying background and foreground noise in \cref{fig:mutual-information}.
Motivating our choice of a model-dependent statistic describing the task difficulty due to spurious correlation, mutual information is unable to account for the effects of foreground and background noise. 

\section{\subpopbench Evaluation Details}
\label{sec:app-subpopbench-evaluation}

To evaluate the validity of common benchmarks, we build upon the comprehensive benchmarking library, \subpopbench, of \citet{yang2023change}.
The worst-group test accuracies we report follow the exact methodology specified by \citet{yang2023change}.
Reported results are the worst-group test accuracy on the test set.
Group attributes are available at all times. 
Following \citet{yang2023change}, we search 16 random hyperparameter configurations, and train 3 random seeds of the best performing configuration, where the ``best'' is the model with the highest worst-group test accuracy on the validation set. 
Vision benchmarks used a ResNet-50 model \citep{he2016resnet}, pretrained on ImageNet-1k \citep{russakovsky2015imagenet}. Language benchmarks used a BERT base uncased model \citep{devlin2019bert}.
For full details see \citet{yang2023change}.

\section{Benchmarks}
\label{sec:app-benchmark-datasets}

\begin{itemize}
\item \waterbirds{} \citep{sagawa2020gdro} is a binary image classification task of land-dwelling vs. water-dwelling birds, where the spuriously correlated attribute is water and land background scenes. 
\item \celeba{} \citep{liu2015celeba} is a binary image classification task of blond vs. not-blond hair color, where the spuriously correlated attribute is annotator-perceived binary gender. \textbf{When training on \celeba, the only output models are capable of is a blond vs. not-blond binary judgment.}
\item ImageNet Backgrounds Challenge, \imagenetbg{} (a.k.a. IN-9L Original; \citealp{xiao2021imagenetbg}) is a multiclass image classification task, where images are selected from 9 coarse classes of ImageNet \citep{deng2009imagenet}. \citeauthor{xiao2021imagenetbg} intend that the background is correlated with the classes, but there is no attribute information available in the \subpopbench formulation. 
Following \citet{yang2023change}, mitigation methods fall back to class labels in the absence of attribute information.
\item \metashift (Indoor/Outdoor Cat vs. Dog; \citealp{liang2022metashift}) is a binary image classification task where the spuriously correlated attribute is indoor or outdoor scenes.
\item \nicopp{} \citep{zhang2023nico} is a multiclass image classification task over common objects, plants and animals where the spuriously correlated attribute is one of (autumn, dim, grass, outdoor, rock, water).
\item \chexpert{} \citep{irvin2019chexpert} is a medical image classification task over chest x-rays, with a binary classification into "finding" or "no finding". The spurious correlated attributes are patient race and gender, following \citep{yang2023change}.
\item \civilcomments{} \citep{borkan2019civilcomments} is a binary text classification task of internet comments to be classified as toxic / not toxic. The spuriously correlated attribute is one of 9 group identities that are the target of the toxicity.
\item \multinli{} \citep{williams2018multinli} is a natural language inference task, comprising sets of sentences where sentences can either entail, contradict or be neutral with one another. The spuriously correlated attribute is the presence of negation words.
\item \dollarstreet{} \citep{rojas2022dollarstreet} is a geographically diverse collection of household images, which in our work we frame as a multiclass object classification task, where the attribute information is geographic region. Although the original \dollarstreet{} more classes, we filter them to only include classes that are present in every region, following \citep{rojas2022dollarstreet}. \textbf{When training on \dollarstreet, models are only capable of outputting one of 42 classes, none of which are related to people.}
\end{itemize}

\subsection{\citybirds}
\citybirds is a new variant of \waterbirds where the confounding background scenes are urban or rural environments.
We generate \citybirds using the \waterbirds generation scripts of \citet{sagawa2020gdro}, modifying the choice of background images.
Background images are drawn from the \textit{Places365} dataset \citep{lopezcifuentes2020places}, with urban backgrounds from the ``street'' and ``downtown'' classes, and rural backgrounds from the ``farm'' and ``field/cultivated'' classes.
The number of samples per group is matched between \waterbirds and \citybirds.

\subsection{Animals vs. Plants (AvP)}
\animalsvsplants is a new binary classification task over diverse images of animals and plants, drawn from two geographic regions, Asia and Europe. The target label is animal or plant, and the group attribute is Asia or Europe.
We construct \animalsvsplants by sampling images from the \textit{GeoYFCC} dataset \citep{dubey2021geoyfcc} of natural images,
limiting our sample to only include classes that are within the hierarchy of the ``animal.n.01'' and ``plant.n.02'' ImageNet hierarchy.
We explicitly exclude all images containing people. 
To identify region as either Asia or Europe, we map the country information provided by GeoYFCC to continents using the \textit{pycountry\_convert} Python package.
The number of samples per group exactly matches \waterbirds.
We validated that geographic information is harder to extract than the animals vs. plants classification task by comparing two linear models, one to predict class membership and another to predict geographic region, over representations extracted from a ResNet-50 
\citep{he2016resnet} pre-trained on ImageNet-1K \citep{russakovsky2015imagenet}.
As expected, geographic performance was at chance level compared to much stronger performance for the target task.

\begin{figure*}[h!]
    \captionsetup[subfigure]{justification=centering}
    \centering
    \begin{subfigure}[t]{0.22\textwidth}
        \centering
        \includegraphics[trim={0 0 0 0},clip,width=\textwidth]{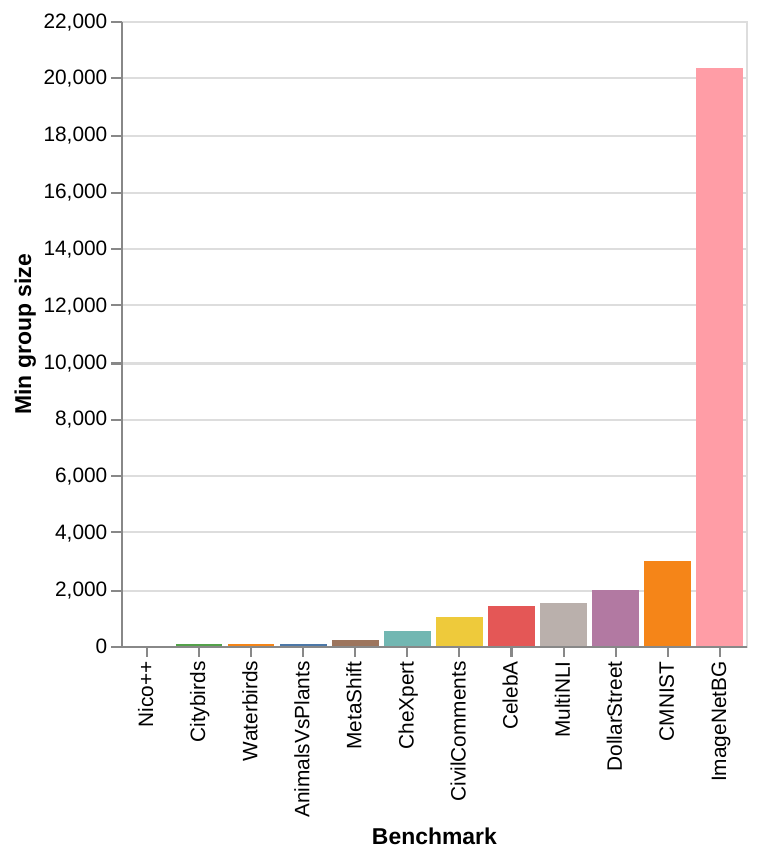}
        \caption{Min group (train)}
    \end{subfigure}
    \begin{subfigure}[t]{0.22\textwidth}
        \centering
        \includegraphics[trim={0 0 0 0},clip,width=\textwidth]{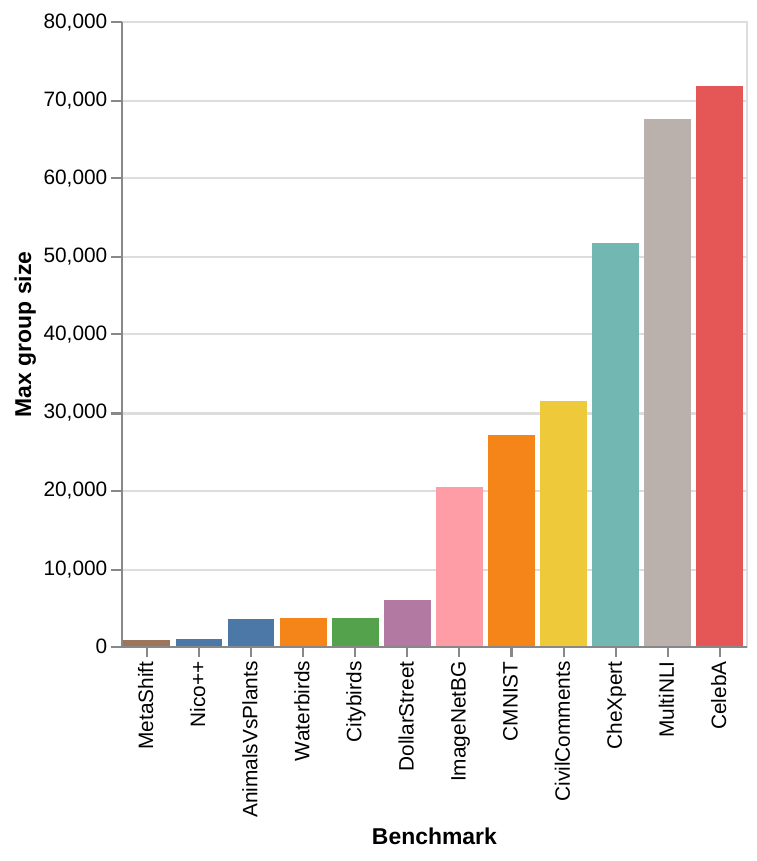}
        \caption{Max group (train)}
    \end{subfigure}
    \hfill
    \begin{subfigure}[t]{0.22\textwidth}
        \centering
        \includegraphics[trim={0 0 0 0},clip,width=\textwidth]{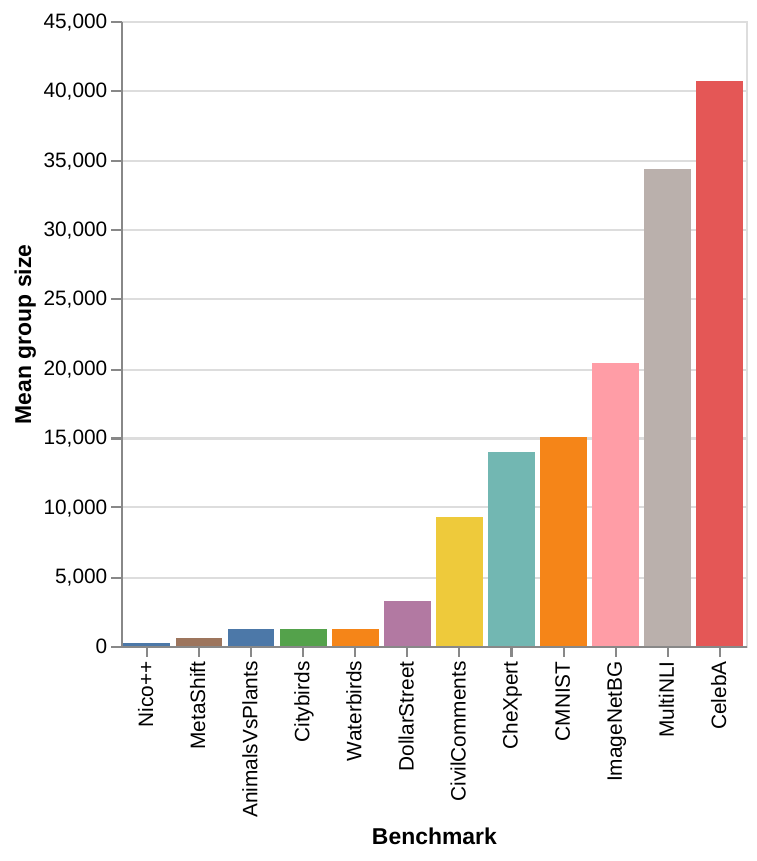}
        \caption{Mean group (train)}
    \end{subfigure}
    \hfill
    \begin{subfigure}[t]{0.29\textwidth}
        \centering
        \includegraphics[trim={0 0 0 0},clip,width=\textwidth]{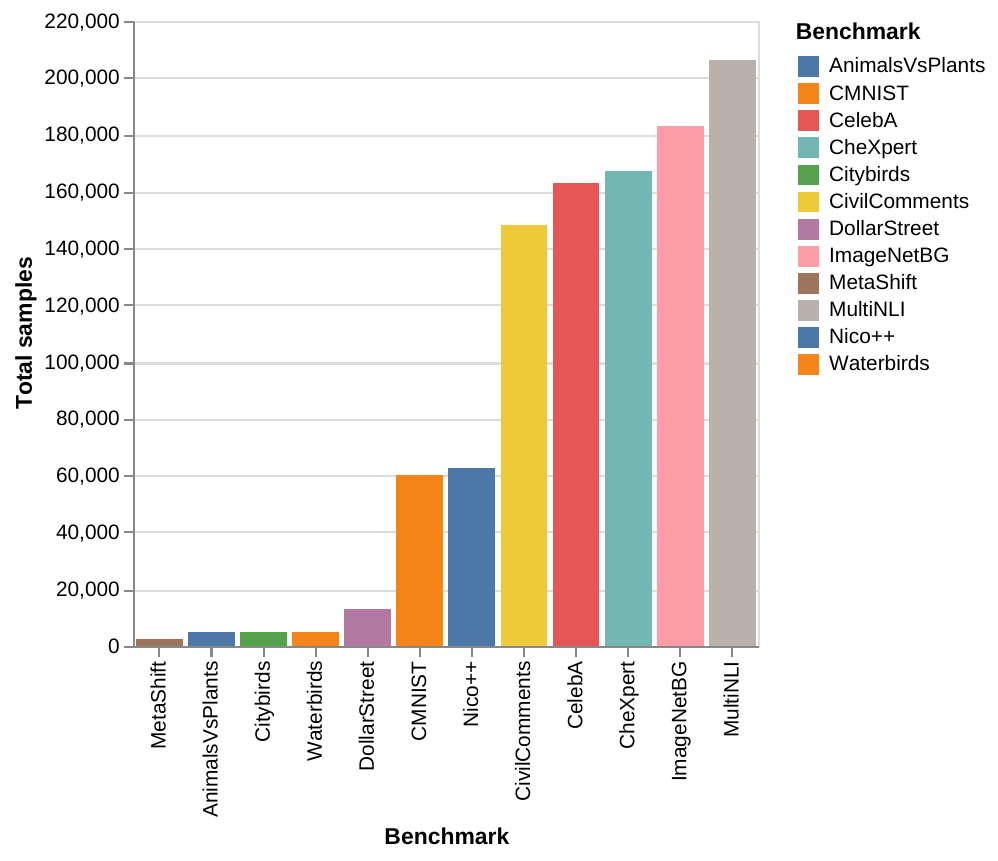}
        \caption{Total samples (train)}
    \end{subfigure}

    \begin{subfigure}[t]{0.22\textwidth}
        \centering
        \includegraphics[trim={0 0 0 0},clip,width=\textwidth]{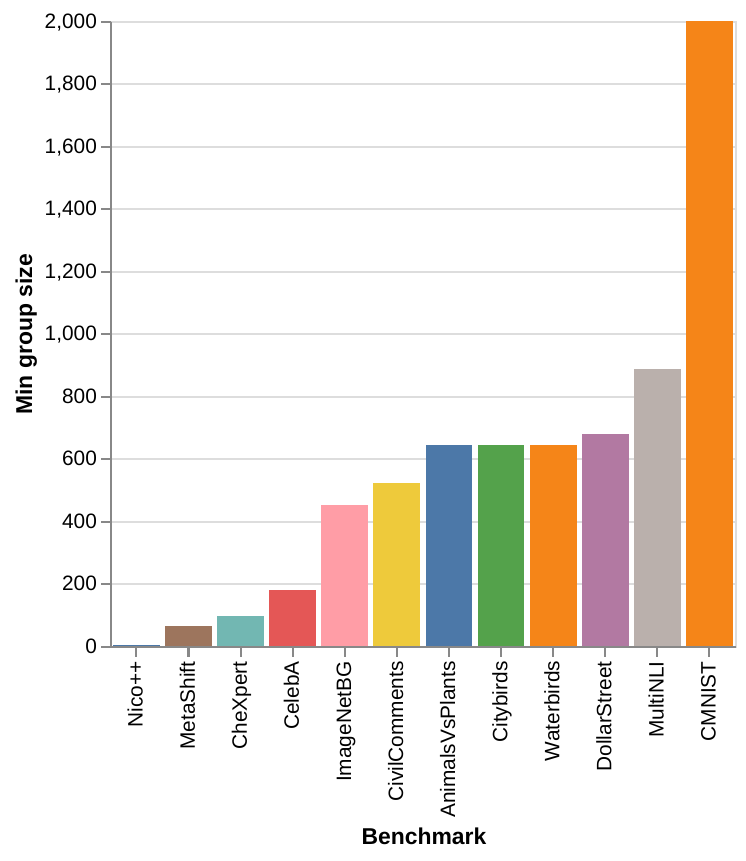}
        \caption{Min group (test)}
    \end{subfigure}
    \begin{subfigure}[t]{0.22\textwidth}
        \centering
        \includegraphics[trim={0 0 0 0},clip,width=\textwidth]{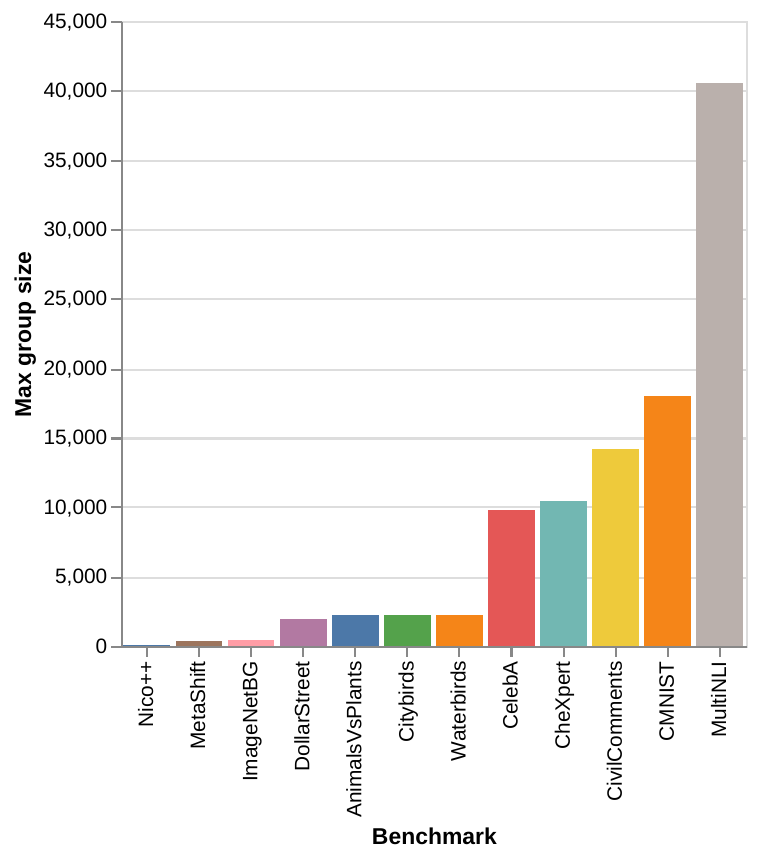}
        \caption{Max group (test)}
    \end{subfigure}
    \hfill
    \begin{subfigure}[t]{0.22\textwidth}
        \centering
        \includegraphics[trim={0 0 0 0},clip,width=\textwidth]{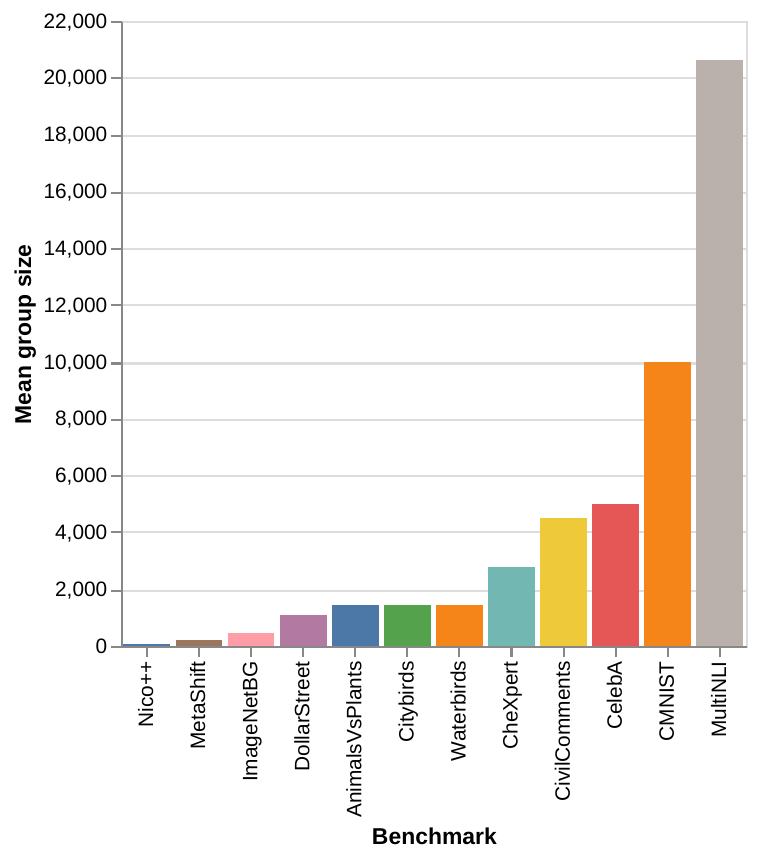}
        \caption{Mean group (test)}
    \end{subfigure}
    \hfill
    \begin{subfigure}[t]{0.29\textwidth}
        \centering
        \includegraphics[trim={0 0 0 0},clip,width=\textwidth]{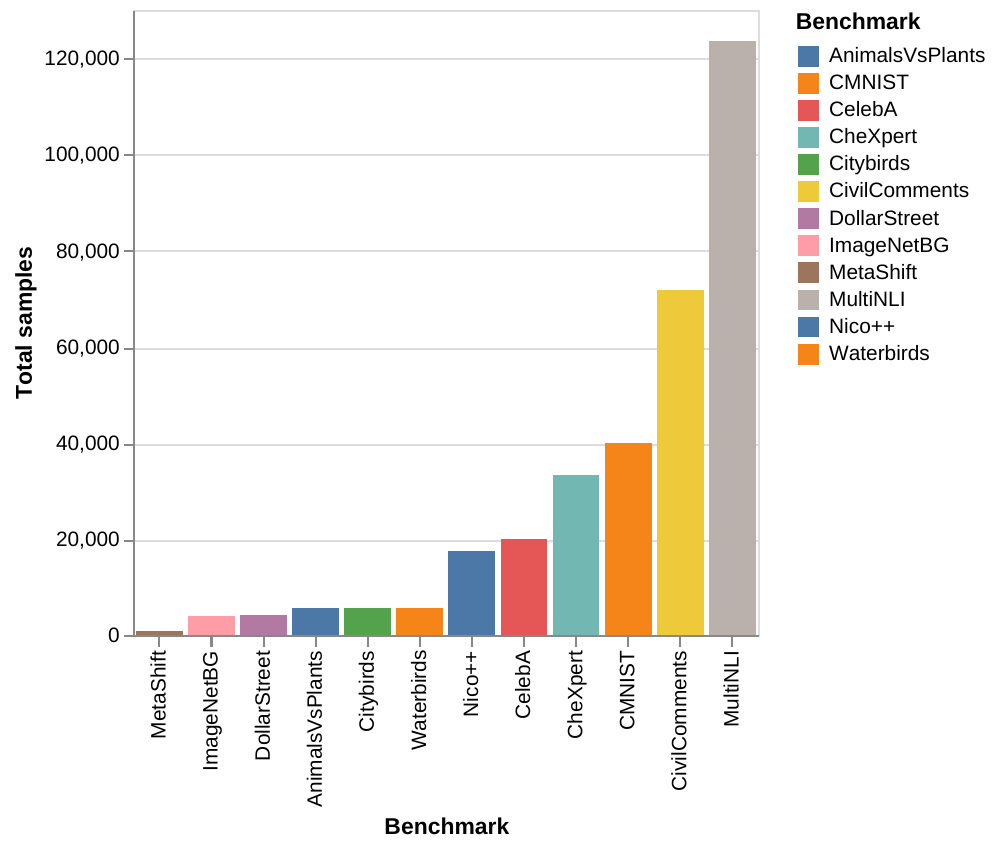}
        \caption{Total samples (test)}
    \end{subfigure}
    \caption{\textbf{(a)} Minimum group size per train dataset; \textbf{(b)} Maximum train group size; \textbf{(c)} Mean train group size; \textbf{(d)} Total number of train samples. \textbf{(e--h)} As above but for test set.}
    \label{fig:num-samples-per-dataset}
\end{figure*}

See \cref{fig:num-samples-per-dataset} for the number of samples per group in each dataset.

\section{\texorpdfstring{$K$}{K} Validation Benchmarks}
\label{sec:app-validating-k}

We implement a number of synthetic modifications to existing benchmarks in order to sanity check our measure $K$.
To test for the effect of the amount of attribute--target correlation, we modify Waterbirds and CelebA.
For Waterbirds, use \citet{sagawa2020gdro}'s generation scripts, varying the ``confounder strength'' argument. 
For CelebA, we use subsampling to balance out the effect of the hair color and gender correlation, while keeping the total number of samples constant.

We also test version of Waterbirds with various degrees of RGB Gaussian noise applied to the background images and the foreground images, though we note that the outline of the bird (if not its detail) is still visible even under high noise.
Given that noise can easily be memorized by contemporary neural networks, we also produce a version with a solid gray background, finding equivalent results to the 100\% noise background case. 
Finally, we test for the effect of attribute noise by randomly flipping a proportion of the annotations, finding the $K$ becomes less able to detect the spurious correlation as attribute noise increases.

\section{\texorpdfstring{$K$}{K} Implementation Details}
\label{sec:app-k-implementation}

To calculate $K$, we train two models, $M_{ERM}$ and $M_{RW}$.
$M_{ERM}$ is a base model (see \cref{sec:app-subpopbench-evaluation}) trained to minimize cross-entropy loss using Adam \citep{kingma2017adam} with learning rate $1e-3$ until validation loss convergence. 
$M_{RW}$ is trained in the same way, though using a reweighted loss function, where the weight of each sample is proportional to the number of samples in each group.

Let $N^{\mathrm{train}}$ be the number of samples in the training set, and $N^{\mathrm{train}}_{a_i,y_i}$ be the number of samples where the attribute $a_i$ and target $y_i$ match the current sample $x_i$, then the weight for sample $x_i$ is 
\begin{equation}
    w_i = \frac{N^{\mathrm{train}}}{N^{\mathrm{train}}_{a_i,y_i}} \ .
\end{equation}

\begin{figure*}[ht!]
    \captionsetup[subfigure]{justification=centering}
    \centering
    \begin{subfigure}[t]{0.16\textwidth}
        \centering
        \includegraphics[trim={0 0 0 0},clip,width=\textwidth]{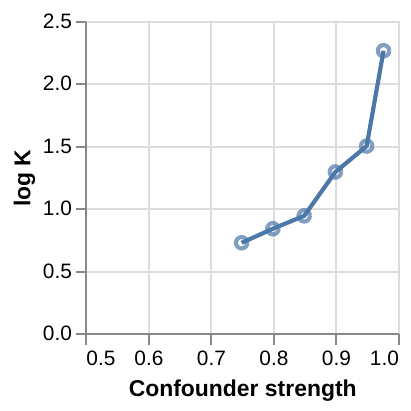}
        \caption{\celeba\\confounder strength}
    \end{subfigure}
    \begin{subfigure}[t]{0.16\textwidth}
        \centering
        \includegraphics[trim={0 0 0 0},clip,width=\textwidth]{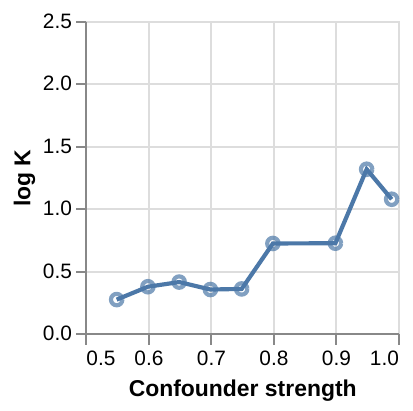}
        \caption{WB confounder strength}
    \end{subfigure}
    \hfill
    \begin{subfigure}[t]{0.16\textwidth}
        \centering
        \includegraphics[trim={0 0 0 0},clip,width=\textwidth]{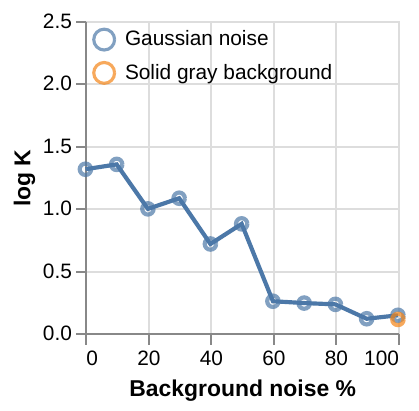}
        \caption{WB\\background noise}
    \end{subfigure}
    \hfill
    \begin{subfigure}[t]{0.16\textwidth}
        \centering
        \includegraphics[trim={0 0 0 0},clip,width=\textwidth]{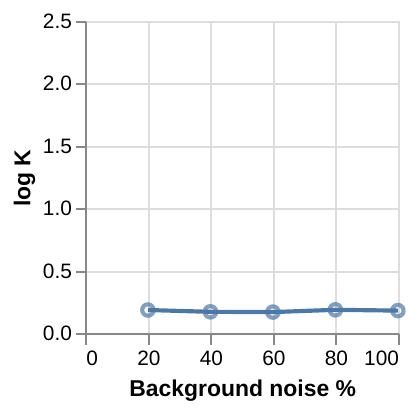}
        \caption{WB background noise, no correlation}
    \end{subfigure}
    \hfill
    \begin{subfigure}[t]{0.16\textwidth}
        \centering
        \includegraphics[trim={0 0 0 0},clip,width=\textwidth]{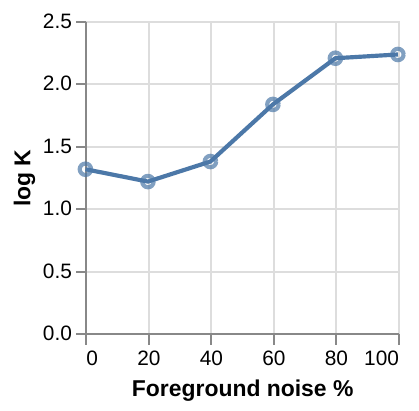}
        \caption{WB foreground noise}
    \end{subfigure}
    \hfill
    \begin{subfigure}[t]{0.16\textwidth}
        \centering
        \includegraphics[trim={0 0 0 0},clip,width=\textwidth]{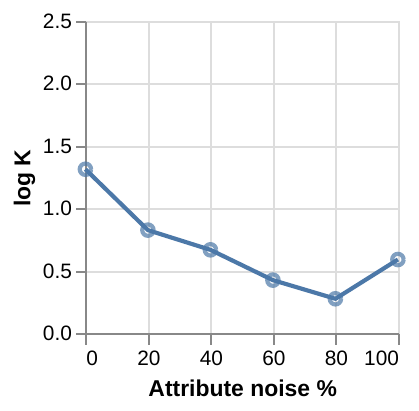}
        \caption{WB attribute noise}
    \end{subfigure}
    \caption{Task difficulty due to spurious correlation, as measured by Bayes Factor $K$ computed using a gDRO reference rather than a reweighted loss reference, for datasets with various synthetic modifications. Increasing the label-attribute correlation (a, b) and foreground noise (e) increases $K$, while increasing background noise (c) or applying a solid gray background (c, orange point) decreases $K$, except in the case where there is no correlation (d). Attribute noise degrades the efficacy of $K$ (f). \textbf{Trends are consistent with \cref{fig:bayes-factor}.}}
    \label{fig:bayes-factor-gdro}
\end{figure*}

\begin{figure*}[ht!]
    \centering
    \includegraphics[trim={0 0 0 0},clip,width=0.32\textwidth]{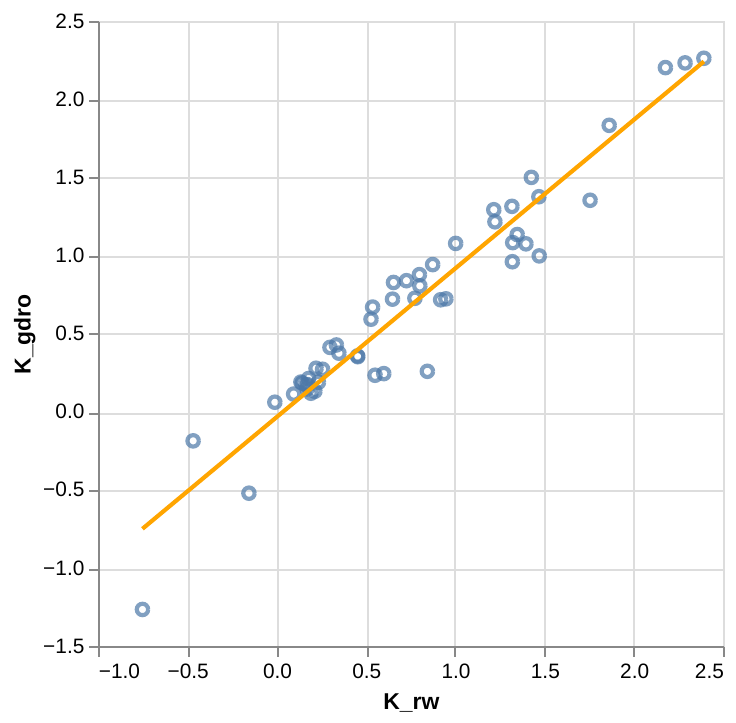}
    \caption{Recomputing Bayes Factor $K$ using a gDRO reference, rather than a reweighted loss reference, does not meaningfully change $K$. \textbf{$K$ is robust to choice of reference model.} (Pearson's $r$ = 0.97, $p < 1e-6$).}
    \label{fig:k-gdro-comparison}
\end{figure*}

Given that our approach is to compare a model using the spurious correlation with one penalized for doing so, it is possible to implement $K$ with an alternative reference method.
We evaluate the robustness of our approach to choice of reference method by computing $K_{\textit{GroupDRO}}$ as follows:
\begin{align*}
    K_{\textit{GroupDRO}} = \log \frac{P(Y^{\mathrm{WG}}_{\mathrm{test}}|X^{\mathrm{WG}}_{\mathrm{test}},M_{\textit{GroupDRO}})}{P(Y^{\mathrm{WG}}_{\mathrm{test}}|X^{\mathrm{WG}}_{\mathrm{test}},M_{ERM})} \ ,
\end{align*}
where $M_{\textit{GroupDRO}}$ is a model trained exactly as above but using GroupDRO \citep{sagawa2020gdro} instead of loss function reweighting. 
\Cref{fig:bayes-factor-gdro} shows that $K_{\textit{GroupDRO}}$ exhibits similar trends to those of $K_{RW}$ reported in \cref{fig:bayes-factor}.
In \cref{fig:k-gdro-comparison}, over all benchmarks and synthetic modifications considered in our paper, we see an almost perfect correlation between $K_{RW}$ and $K_{\textit{GroupDRO}}$.
Our results indicate that our measure of task difficulty due to spurious correlation is not senstive to choice of reference method.

\begin{figure*}[ht!]
    \captionsetup[subfigure]{justification=centering}
    \centering
    \hfill
    \begin{subfigure}[t]{0.32\textwidth}
        \centering
        \includegraphics[trim={0 0 0 0},clip,width=\textwidth]{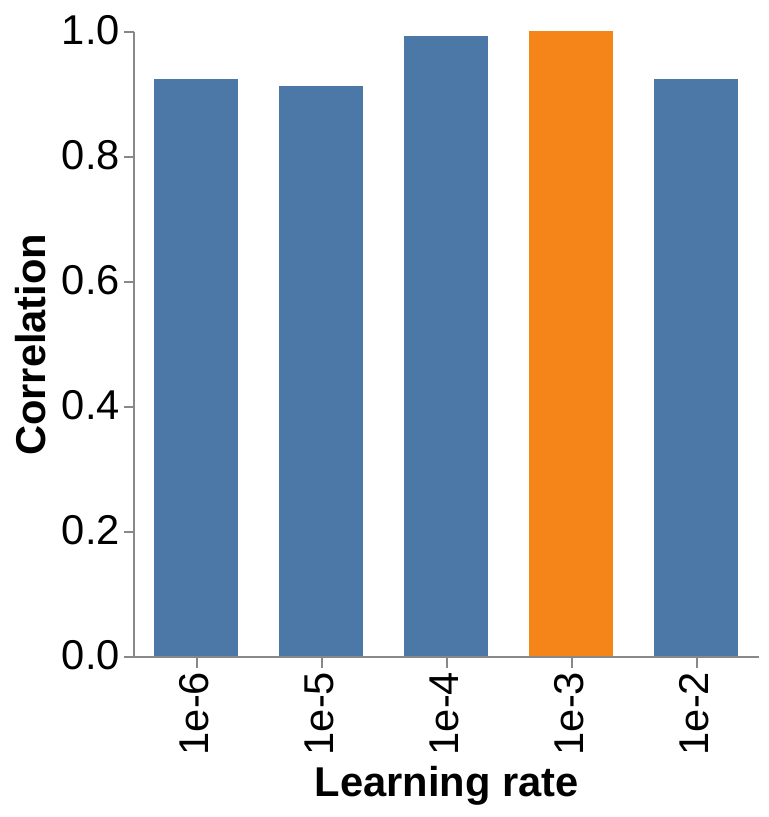}
        \caption{Learning rate}
    \end{subfigure}
    \hfill
    \begin{subfigure}[t]{0.32\textwidth}
        \centering
        \includegraphics[trim={0 0 0 0},clip,width=\textwidth]{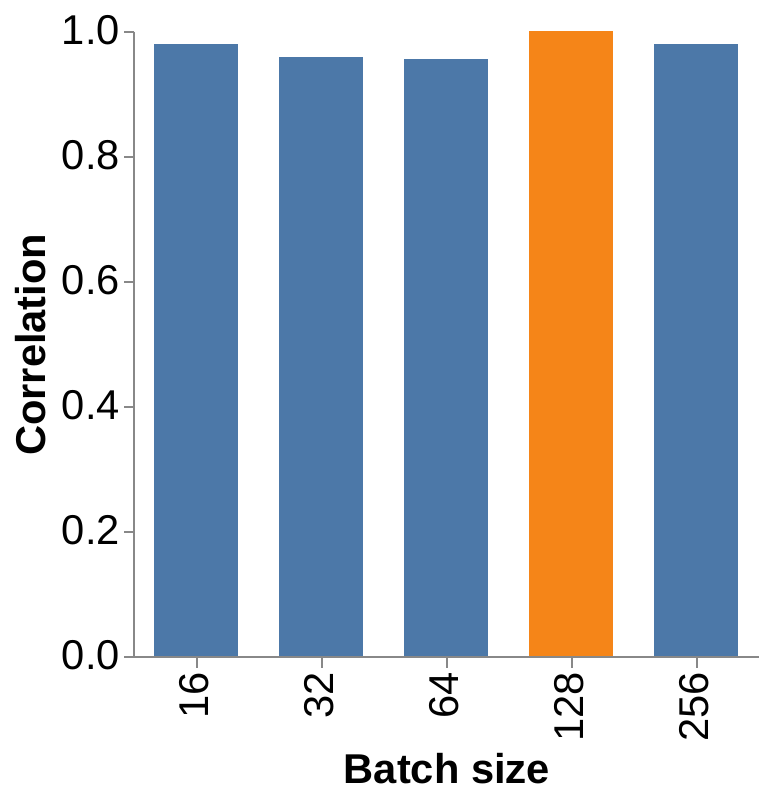}
        \caption{Batch size}
    \end{subfigure}
    \hfill
    \caption{Correlation (Pearson's $r$) between $K$ as reported in main text (orange bar) and variants $K$ calculated with different \textbf{(a)} learning rates and \textbf(b) batch sizes, over benchmarks reported in \cref{table:predicted-method}. Training $M_{ERM}$ and $M_{RW}$ with other learning rates or batch sizes produces a $K$ that is consistently significantly highly positively correlated ($r > 0.9$; $p \leq 0.001$) with the original $K$ across all settings considered. }
    \label{fig:bayes-factor-hyperparams}
\end{figure*}

A natural consequence of a model-dependent metric is a degree of sensitivity to the optimization procedure used when training the underlying model.
For example, it is possible that $K$ may vary according to the hyperparameters chosen for training $M_{ERM}$ and $M_{RW}$.
To evaluate this, we recalculate $K$ under different learning rates and batch sizes, for each of the benchmarks reported in \cref{table:predicted-method}, and evaluate whether the resulting sets of $K$ are consistent with $K$ as originally reported.
We vary hyperparameters in lockstep between the two models, such that the learning rate for both models is always equal, as is the batch size, resulting in a like-versus-like, rather than best-versus-best comparison. 
Note that for the batch size experiments, we exclude \nicopp and \dollarstreet due to training instability resulting from unsuitable batch sizes.
In \cref{fig:bayes-factor-hyperparams}, we present the correlation (Pearson's $r$) of the resulting $K$ versus the reference implementation, finding that regardless of hyperparameter choice, $K$ appears to produce a significantly highly positively correlated set of $K$ ($r > 0.9$; $p \leq 0.001$).
Thus, we conclude that $K$ is practically robust in light of reasonable hyperparameter optimization. 

\section{Evaluating Method Sensitivity}
\label{sec:app-method-sensitivity}
We test whether the variability in method performance is a function of varying $K$ by fitting a linear model using OLS to the worst-group test accuracies and the benchmark's $K$.
We evaluate sensitivity in terms of the proportion of variance explained, using the linear model's Coefficient of Determination, $R^2$. 
A higher $R^2$ indicate that $K$ explains more of the variance in method performance, indicating greater sensitivity.

\end{document}